# One-shot learning of paired association navigation with biologically plausible schemas

M Ganesh Kumar[1,2,3,4], Cheston Tan[5], Camilo Libedinsky[1,2,6,7,12, 13], Shih-Cheng Yen[1,2,3,13*], Andrew Y. Y. Tan[8,9,10,11,12,13*]

1. Integrative Sciences and Engineering Programme, NUS Graduate School, National University of Singapore
2. The N.1 Institute for Health, National University of Singapore
3. Engineering Design and Innovation Centre, College of Design and Engineering, National University of Singapore
4. School of Engineering and Applied Sciences, Harvard University
5. Institute for Infocomm Research, Agency for Science, Technology and Research
6. Department of Psychology, National University of Singapore
7. Institute of Molecular and Cell Biology, Agency for Science, Technology and Research
8. Department of Physiology, Yong Loo Lin School of Medicine, National University of Singapore
9. Neurobiology Translational Research Programme, Yong Loo Lin School of Medicine, National University of Singapore
10. Healthy Longevity Translational Research Programme, Yong Loo Lin School of Medicine, National University of Singapore
11. Cardiovascular Disease Translational Research Programme, Yong Loo Lin School of Medicine, National University of Singapore
12. Neurobiology Programme, Life Sciences Institute, National University of Singapore
13. NUS Artificial Intelligence Institute, National University of Singapore

* Contributed equally to this work.

Correspondence: M Ganesh Kumar (m_ganeshkumar@u.nus.edu), Andrew Tan (atyy@alum.mit.edu)

# Abstract

After learning a multiple paired-association (MPA) navigation task, rodents can rapidly learn new cue–location associations after a single exposure, a phenomenon proposed to be facilitated by cognitive schemas. However, it remains unclear what schemas support this behavior and how they could be implemented in biological neural circuits. Here, we identify a set of schemas sufficient for gradual then one-shot paired-association navigation, and implement them in a fully neural agent using biologically plausible plasticity rules. The agent gradually acquires a low-dimensional spatial metric representation using a path-integration based temporal-difference error, enabling self-localization across different environments. It also contains a memory network that forms one-shot associations between sensory cues and goal locations, implemented using reward-modulated exploratory Hebbian (EH) plasticity. A third network performs vector subtraction between the agent's current and goal location to compute goal-directed actions. Supplementing these schemas with an actor–critic enables one-shot navigation past obstacles. Lastly, learning a working-memory gating mechanism using the temporal-difference error enables one-shot learning even with brief cue and distractor presentation. The model reproduces key behavioral features of rodent learning in the MPA task and generates testable predictions about the neural mechanisms underlying schema-based one-shot learning.

# Significance Statement

Schema-dependent learning is exemplified by better recall of a lecture when one already has a schema for understanding. A rodent navigation task showing a shift from slow to rapid learning has been proposed to reflect such schema use, yet how schemas are neurally implemented and enable faster learning is poorly understood. We clarify this gap in this computational study by describing several schemas and implement them as neural networks with biologically plausible learning rules. Composing them into an agent reproduces schema-dependent rodent learning behavior. Because schemas also shape self beliefs, clarifying their neural basis has implications for psychopathology. At the same time, understanding how biological schemas support rapid learning may inform advances in machine learning.

# Introduction

Schemas are knowledge structures or frameworks that describe relationships among stimuli, actions, and outcomes, enabling efficient learning and generalization (Rumelhart, 1980; Rumelhart & Ortony, 1977). Importantly, by organizing prior experience, schemas can aid rapid learning (Bartlett & Burt, 1932). For example, one better recalls lecture content in a subject for which one already has a framework for understanding. Beyond adaptive learning, maladaptively formed schemas have been implicated in psychiatric disorders including depression where persistent negative schemas about self are thought to bias perception, memory, and action selection (Bentall et al., 2001; Czekalla et al., 2026; Gagne et al., 2020). Hence, understanding how schemas are acquired, utilized, and adapted has implications for cognition and clinical intervention. Despite this relevance, how schemas are implemented by biological neural circuits, synaptic plasticity rules, and neuromodulatory signals, remains poorly understood.

Spatial navigation is a well-studied domain in which schemas have been proposed to support rapid learning. In a seminal set of experiments, Tse and colleagues showed that rodents performing a two-stage multiple paired-association (MPA) task exhibit a striking transition from slow learning to one-shot learning (Tse et al., 2007). In the task, rodents navigate an open-field environment where sensory cues presented at trial onset indicates the location of a corresponding hidden reward. In the first stage, animals gradually learned six cue–location associations. In the second stage, new cues paired with new locations were introduced. Surprisingly, rodents needed only a single exposure to each new cue to navigate to the correct location on subsequently encountering the cue. Lesion and inactivation studies implicated hippocampal and prefrontal cortex, suggesting that a task structure or schema acquired during the first stage enabled the one-shot learning in the second stage (Tse et al., 2007, 2011). However, the computational purpose of this schema and its neural implementation remain unresolved.

Related phenomena have been studied in machine learning on transfer or meta-learning, where prior learning accelerates subsequent learning to being few-shot or one-shot (Finn et al., 2017; Hospedales et al., 2021; Lee et al., 2023; Lin et al., 2025; J. X. Wang et al., 2018). Although such models can reproduce rapid learning in variants of the MPA task (Foster et al., 2000; Hwu & Krichmar, 2020; McClelland, 2013; Whittington et al., 2020), they typically rely on symbolic algorithms or biologically implausible learning rules such as backpropagation or contrastive Hebbian learning that are difficult to reconcile with known properties of biological synaptic plasticity, including locality, causality, and neuromodulatory control (Bellec et al., 2020; Lillicrap et al., 2020; Movellan, 1991; Murray, 2019; Overwiening et al., 2025; Rumelhart et al., 1986; J. Wang et al., 2025). Moreover, these approaches do not explicitly identify the schemas required for gradual structure learning and subsequent one-shot association.

Thus, a gap remains: what schemas support one-shot learning in spatial navigation, and how are these schemas learned and implemented in biological neural circuits? Addressing this question requires moving beyond abstract descriptions towards explicit formulations and neurally plausible mechanisms that bridge conceptual, algorithmic and implementation levels.

Here, we address these issues by identifying a set of schemas sufficient for paired-association navigation and implementing them in a fully neural agent with biologically plausible learning rules. We first describe navigation-relevant schemas at the computational level, including spatial localization, cue–goal association, and goal-directed action selection, and relate each to corresponding algorithmic and neural implementations. We then compose these schemas into a single agent whose synapses are updated by local, reward-modulated learning rules, enabling initial gradual learning and subsequent one-shot learning of new cue–location associations. Next, we show that supplementing the schema-based agent with an actor–critic allows one-shot navigation in environments containing obstacles. Finally, we demonstrate that when target cues are presented alongside distractors, gradual temporal-difference learning of a working-memory gating mechanism enables generalization. Together, these results provide a biologically plausible account of schema-based one-shot navigation and yield testable predictions about the neural circuits and learning signals underlying this behavior.

# Methods

## Algorithmic implementations of schemas

We build on work by Foster and colleagues (2000), who demonstrated a partly biologically-plausible agent that modelled rodent one-shot learning of a delayed matching-to-place (DMP) task (Steele & Morris, 1999). Foster and colleagues' agent may be thought of as composed of 3 schemas: (1) LEARN METRIC REPRESENTATION, (2) LEARN GOAL COORDINATES, (3) NAVIGATE. Here we provide details of how these three schemas can be implemented algorithmically, and using neural circuits and biologically plausible learning rules. We first describe the general neuron model, then each schema in turn together with the inputs and networks it uses, followed by model-free reinforcement learning, the working memory gating mechanism, and task descriptions.

### General neuron model

The membrane potential dynamics $x_j(t)$ of all neurons, except place cells and units within the neural NAVIGATE schema, were simulated as

$$\tau \dot{x}_j(t) = -x_j(t) + \sum_{i=1}^{N} W_{ji} u_i(t) + \sqrt{\tau\sigma^2}\xi_j(t)\,, \tag{1}$$

with membrane time constant $\tau = 100\ ms$. Inputs $u_i(t)$ were linearly weighted using synaptic weights $W_{ij}$, and stochasticity defined using Gaussian white noise process $\xi_j(t)$ with zero mean and variance scaled by $\sigma$. The firing rates were modelled using either a linear or nonlinear activation function. Each neuron's dynamics was discretized with the Euler–Maruyama method:

$$x_j(t) = (1-\alpha)x_j(t-\Delta t) + \alpha\left(\sum_{i=1}^{N} W_{ji} u_i(t)\right), \tag{2}$$

where $\alpha \equiv \frac{\Delta t}{\tau}$ and $N(0,1)$ is the standard normal distribution. We used a time step of $20\ ms$ for all simulations. The specific implementation is outlined in the subsequent sections.

### (1) LEARN METRIC REPRESENTATION algorithm

*Place cells*

All agents have 49 place cells whose firing rates depend on the agent's position in the arena $s(t)$. The firing rate of the $i$th place cell is:

$$u_i^{pc}(t) = exp\left(-\frac{(s(t)-s_i)^2}{2\sigma_{pc}^2}\right), \tag{3}$$

with width $\sigma_{pc} = 0.267\ m$, and place cell centers $s_i$ spaced $0.267\ m$ apart at the intersections of a regular 7-by-7 grid. Place cell selectivity was shuffled by randomly sampling $s_i$ from the grid space.

*Learning the metric representation*

This schema allows the agent to self-localize in an environment by learning coordinates that provide a continuous and low-dimensional metric representation. An agent can estimate its current position $\boldsymbol{p}(t)$ based on the firing rate of the X and Y coordinate cells which follow the dynamics of the membrane potential

$$\tau\dot{p}_j(t) = -p_j(t) + \sum_{i=1}^{P} W_{ji}^{MR} u_i{}^{pc}(t) + \sqrt{\tau\sigma_{coord}^2}\xi_j(t)\,, \tag{4}$$

where $\sigma_{coord}^2 = 10^{-8}$ and $W_{ji}^{MR}$ is the synaptic weights from place cells to coordinate cells. Learning a metric representation is formulated as a path integration learning problem by integrating place cell activity and self-motion information $\hat{\boldsymbol{a}}(t)$. The self-motion information is the actual displacement of the agent in an environment after correcting for boundary conditions, making it different from the action $\boldsymbol{a}(t)$ specified by the agent (Eq. 32, below). An agent can estimate its current coordinates by vector addition between its displacement and the previously estimated coordinates

$$p_j(t) = p_j(t - \Delta t) + \hat{a}_j(t)\,, \tag{5}$$

which yields the self-consistency equation

$$p_j(t) - p_j(t - \Delta t) - \hat{a}_j(t) \equiv 0\,, \tag{6}$$

if the agent performed perfect path integration to accurately estimate its current coordinates. Errors arising from coordinate estimation can be converted into a temporal difference error

$$\delta_j^{MR}(t) = p_j(t) - p_j(t - \Delta t) - \hat{a}_j(t)\,, \tag{7}$$

which the agent minimizes by first computing an eligibility trace of the place cell activity

$$\tau_{MR}\dot{e}_i(t) = -e_i(t) + u_i{}^{pc}(t)\,, \tag{8}$$

with $\tau_{MR} = 1000\ ms$ and using the two-factor Hebbian rule by taking the presynaptic place cell activity modulated by the path integration derived TD error

$$\Delta W_{ji}^{MR}(t) = \eta_{MR} \cdot e_i(t) \cdot \delta_j^{MR}(t)\,, \tag{9}$$

with $\eta_{MR} = 0.01$. Reducing the time constant $\tau_{MR}$ to $200\ ms$ and $100\ ms$ allowed the agent to learn the metric representation though convergence was increasingly slower (See Figure S1A). While Foster et al. (2000) proposed the schema, whether the learned metric representation could transfer to a new environment was not previously studied.

## (2) LEARN FLAVOR-LOCATION algorithm

*Sensory cues*

Each cue is encoded by a one-hot vector $u^{cue}$ of length 18 with gain 3, for example, $u^{cue} = [0,3,0,\ldots]$ for the second cue. The cue was constant throughout each trial, except for the working memory task in Figure 7 where the cue was presented 1 second after the start of each trial for 2 seconds, similar to the experiment by Tse et al. (2007). During the cue presentation period, place cell activity and agent actions were silenced to simulate cue presentation to the rat in the starting box with no knowledge of its position in the maze. Subsequently, $u^{cue}$ was set to zero while place cell activity and agent actions were switched on for navigation.

*Feedforward and reservoir networks*

Following Kumar et al. (2022), place cell activities and sensory cue were concatenated to form an input vector

$$\boldsymbol{u}(t) = [\boldsymbol{u}^{pc}(t), \boldsymbol{u}^{cue}(t)\,], \tag{10}$$

of length $N_{inputs} = 67$ and passed to a reservoir of $N = 1024$ recurrently connected neurons as inputs. The firing rates of the reservoir neurons are

$$r_j(t) = \phi[x_j(t)], \tag{11}$$

with the nonlinear activation function defined by Kumar et al. (2022)

$$\phi(x) = \begin{cases} 0, & x < 3 \\ x, & x \geq 3 \end{cases}, \tag{12}$$

and membrane potential dynamics

$$\tau \dot{x}_j(t) = -x_j(t) + \sum_{i=1}^{N_{inputs}} W_{ji}^{inp} u_i(t) + \lambda \sum_{k=1}^{N} W_{jk}^{rec} \tanh[x_k(t)] + \sqrt{\tau \sigma_{res}^2}\,\xi_j(t), \tag{13}$$

with $\lambda = 1.5$ and $\sigma_{res} = 0.025$. The synaptic weights $W_{ji}^{inp}$ were drawn from a uniform distribution between $[-1,1]$. ${W_{jk}}^{rec}$ specifies the recurrent synaptic connections drawn from a Gaussian distribution with zero mean and variance $1/pN$ with connection probability $p = 0.1$. A feedforward network implements Eq. 13 with the recurrent term omitted ($\lambda = 0$).

*Associating and deleting cue–goal pairs*

Foster's (2000) agent used a non-neural, symbolic memory implementation that allowed an agent to store a single set of goal coordinates in one shot when a reward was disbursed and deleted the wrong coordinates when the reward was not reached. We view this as a LEARN GOAL COORDINATES schema. We replaced the symbolic memory implementation with the LEARN FLAVOR-LOCATION schema, which is a hetero-associative memory neural network that can learn multiple cue-coordinate paired associations in one shot, and delete specific associations if they are wrong.

When the agent navigates around the arena and receives a reward, its current estimated location $\boldsymbol{p}(t)$ is taken to be the goal coordinates. The symbolic agent contains a key–value association matrix. When the reward value is greater than zero, the cue index $C \in \{1, 2, 3, \dots, 17, 18\}$ is determined using the one-hot encoded cue vector $\boldsymbol{u}^{cue}$ with a nonzero value. The cue index specifies the row in which the vector $\boldsymbol{u}^{cue}$ needs to be stored in the key matrix $K$ and the agent's current coordinates $\boldsymbol{p}(t) = (x, y)$ concatenated with a recall value of 1 is stored in the value matrix. Refer to Fig. 2B for an example. In the subsequent trial, $\boldsymbol{u}^{cue}$ is treated as a query and a distance-based metric

$$A(t) = softmax(\boldsymbol{u}^{cue}(t) \cdot K^T)\,, \tag{14}$$

computes the memory attention weight $A(t)$ which specifies the probability distribution if and where the flavor cue vector $\boldsymbol{u}^{cue}$ is stored in the key matrix. A dot product between the attention weight and the value matrix

$$\boldsymbol{g}(t) = A(t) \cdot V\,, \tag{15}$$

recalls the goal coordinates and recall value $[x, y, 1]$ as the vector $\boldsymbol{g}(t)$. The recall value describes the accuracy of recalling the goal coordinates i.e., when the recall of goal coordinates is imperfect, recall value will be lower than 1. If the trial ends and no reward is disbursed, the entire row of the key and value matrices corresponding to the cue index $C$ is set to 0 to delete the cue and coordinate association.

Instead of using a symbolic key-value matrix, we propose using a neural network to rapidly learn and delete cue-coordinate associations. Three readout units either from a feedforward or reservoir network are trained to recall the X, Y goal coordinates and recall value when it receives flavor cues as inputs. The firing rate of the goal coordinate neurons $g_i(t)$ follows the membrane potential dynamics

$$\tau \dot{g}_i(t) = -g_i(t) + {g_i}^{noisy}(t)\,, \tag{16}$$

with ${g_i}^{noisy}(t)$ given by a vector sum of the reservoir activity and $W_{ij}^{FL}$ synaptic weights

$${g_i}^{noisy}(t) = \sum_{j=1}^{N} W_{ij}^{FL} r_j(t) + \sqrt{\tau \sigma_{FL}^2}\,\xi_i(t)\,, \tag{17}$$

as well as the exploratory white noise with $\sigma_{FL}^2 = 0.05$. To form an association between the flavor cue and the agent's coordinates, a target vector $g_i^*(t)$ is determined according to

$$g_i^*(t) = \left[p_i(t), \Theta[R(t)]\right], \tag{18}$$

which is a concatenation of the agent's estimated coordinates $\boldsymbol{p}(t)$ and the reward value that is transformed using a step function $\Theta$ to be a value of 1 when the reward is positive and a value of zero otherwise (Eq. 24). The synaptic weights are trained either by the reward modulated least mean squares (LMS) delta rule (Maass et al., 2007; Sussillo & Abbott, 2009) which takes the presynaptic reservoir firing activity and the vector error between the target vector $g_i^*(t)$ and goal coordinate neurons $g_i(t)$

$$\Delta W_{ij}^{goal}(t) = \eta_{goal} \cdot r_j(t) \cdot \left(g_i^*(t) - g_i(t)\right) \cdot \Theta[R(t)] \tag{19}$$

or the reward modulated exploratory Hebbian (EH) rule (Hoerzer et al., 2012) that takes the presynaptic reservoir activity and the difference between the noisy and smooth goal neuron firing activity as postsynaptic neurons

$$\Delta W_{ij}^{FL}(t) = \eta_{FL} \cdot r_j(t) \cdot \left(g_i^{noisy}(t) - g_i(t)\right) \cdot M(t) \cdot \Theta[R(t)]\,, \tag{20}$$

A sparse modulatory factor $M(t)$ is computed

$$M(t) = \begin{cases} 1, & \bar{P}(t) < P(t) \\ 0, & otherwise\,, \end{cases} \tag{21}$$

where the performance index $P(t)$ is the negative mean squared error between the target vector $g_i^*(t)$ and the noisy goal neuron $g_i^{noisy}(t)$ activity

$$P(t) = -\sum_{i=1}^{3} \left[g_i^*(t) - g_i^{noisy}(t)\right]^2\,, \tag{22}$$

and a low pass filter of the performance index is $\bar{P}(t)$ given as

$$\tau \frac{d\bar{P}}{dt}(t) = -\bar{P}(t) + P(t)\,, \tag{23}$$

with $\tau = 100\,ms$, the same as the neuronal time constant. The EH rule is considered to be biologically plausible as it uses only local presynaptic and postsynaptic information while the modulatory factor is a sparse scalar value (Hoerzer et al., 2012; Legenstein et al., 2010).

Importantly, the Hebbian plasticity rule needs to be modulated by the presence of the reward so that the coordinates at which a reward is disbursed is learned as the goal coordinates. A step function $\Theta$ is used to transform the reward value

$$\Theta[R(t)] = \begin{cases} 1, & R(t) > 0 \\ 0, & otherwise\,, \end{cases} \tag{24}$$

so that reward modulation is 0 for negative or no rewards or 1 for positive rewards.

To delete spuriously formed cue specific goal association, synaptic depression is modelled using an acetylcholine-modulated Hebbian rule (Zannone et al., 2018) which takes the reservoir activity as the presynaptic activity, the goal readout activity as the postsynaptic activity and a constant negative scalar value as the global modulatory factor

$$\Delta W_{ij}^{FL}(t) = \eta_{FL} \cdot r_j(t) \cdot g_i(t) \cdot -\Omega_{Ach}\,, \tag{25}$$

where a higher valued acetylcholine factor $\Omega_{Ach}$ causes a faster rate of synaptic depression (Fig. 2D). Associations can be either deleted immediately with a large acetylcholine signal or gradually decay with a small constant signal. The value of the acetylcholine factor is optimized according to the task (Table 1).

**(3) NAVIGATE algorithm**

*Goal-vector computation*

Given a set of current and goal coordinates, an agent can perform vector subtraction to determine the direct heading to reach the goal. Foster et al. (2000) had a non-neural, symbolic implementation of this schema. Concretely, vector subtraction is performed between the goal and agent's current coordinates

$$d_{j\in\{x,y\}}(t) = g_{j\in\{x,y\}}(t) - p_j(t)\,, \tag{26}$$

to determine the direction to move towards the goal from an agent's current position. A spatial direction that is closest to the computed vector $d_{j\in\{x,y\}}(t)$ is chosen out of the 40 possible directions defined by the actor (Eq. 32, below)

$$q_k^{NAV}(t) = softmax\left(\sum_{j\in\{x,y\}}^{M} K_{kj}^{actions} d_j\right) \cdot \varepsilon(t)\,, \tag{27}$$

to directly head towards the goal. If the recall value $g_{j=3}(t)$ is less than a pre-set threshold value of 0.6, the output is suppressed

$$\varepsilon(t) = \begin{cases} 1, & 0.6 < g_{j=3}(t) \\ 0, & otherwise \end{cases}\,, \tag{28}$$

otherwise, the direction of movement $q_k^{NAV}(t)$ is passed to the actor (Eq. 30, below) to influence the action $a(t)$.

*Action selection*

All agents have an actor of $M = 40$ neurons with the firing rate of the $k$th actor neuron

$$\rho_k(t) = ReLU[q_k(t)]\,, \tag{29}$$

determined by a rectified linear unit activation (ReLU) function of the membrane potential $q_k$ with dynamics

$$\begin{aligned} \tau\dot{q}_k(t) = -q_k(t) + \beta^{control} q_k^{NAV}(t) + (1-\beta^{control}) \sum_{j=1}^{N} W_{kj}^{actor} r_j(t) \\ + \sum_{h=1}^{M} W_{kh}^{lateral} \rho_h(t) + \sqrt{\tau\sigma_{actor}^2}\,\xi_k(t)\,, \end{aligned} \tag{30}$$

with $\sigma_{actor} = 0.25$. Note $q_k^{NAV}$ is the additional input from the NAVIGATE schema (Eq. 27). The synaptic weights $W_{kj}^{actor}$ linearly weight the reservoir activity. $\beta^{control}$ determines the contributions from the reservoir and the NAVIGATE schema in influencing the agent's actions. $\beta^{control} = 0.3$ means input to the actor was 30% from NAVIGATE and 70% from the reservoir; $\beta^{control} = 0$ is a pure actor–critic agent; $\beta^{control} = 1$ is a pure schema agent.

$\beta^{control} = 0.3, 0.4, 0.9$ were used for the navigation tasks in Figure 4, 6 and 7 respectively after optimization. Refer to Figure S5 for performance curves when varying $\beta^{control}$. $W_{kh}^{lateral}$ is the lateral inhibition defined using

$$W_{kh}^{lateral} = \frac{w_-}{M} + w_+ \frac{f(k,h)}{\sum_h^M f(k,h)}, \tag{31}$$

with $f(k,h) = (1-\delta_{kh})e^{\varphi cos(\theta_k - \theta_h)}$, $w_+ = 1$, $w_- = -1$ and $\varphi = 20$, connect the actor neurons into a ring attractor that smooths the agent's trajectory (Frémaux et al., 2013; Zannone et al., 2018). The $k$th actor neuron represents a spatial direction $\theta_k = 2\pi k/M$ and the action

$$a(t) = \frac{a_0}{M} \sum_k^M \rho_k(t)[sin\,\theta_k, cos\,\theta_k\,], \tag{32}$$

is the vector sum of directions weighted by each actor neuron's firing rate, with $a_0 = 0.03$ translating to the agent moving at about 0.8 $\mathrm{ms}^{-1}$. The lateral inhibition enforces competition amongst actor neurons where similarly tuned neurons excite each other and are active while differently tuned neurons suppress each other and are silent.

*Pretraining the NAVIGATE network*

A dataset with different combinations of current $\boldsymbol{p}(t)$, goal and recall values $\boldsymbol{g}(t)$ as input and the corresponding suppressed or unsuppressed direction of movement $q^{NAV}(t)$ as output was generated using equations 26–28. Different neural network architectures with one to two hidden layers, each with 64 to 1024 hidden units with ReLU activation function were trained using backpropagation for 20 epochs to minimize the mean squared error of the dataset. All the architectures had five input neurons corresponding to the agent's current X and Y coordinates, recalled X and Y goal coordinates and the recall value, and a top layer with 40 neurons with linear activation function corresponding to the direction of movement in Eq. 32. The network that achieved the lowest validation errors while having low computational complexity was chosen to be the NAVIGATE neural schema. This was a network with two hidden layers each with 128 units. The synaptic weights of this network were fixed and used as a static module as the agent learned the DMP and MPA tasks.

## Model-free reinforcement learning

Agents with a critic neuron have neuron firing rate

$$v(t) = ReLU[\varsigma(t)], \tag{33}$$

with membrane potential dynamics

$$\tau\dot{\varsigma}(t) = -\varsigma(t) + \sum_{j=1}^{N} w_j^{critic} r_j(t) + \sqrt{\tau\sigma_{critic}^2}\xi(t), \tag{34}$$

with $\sigma_{critic} = 10^{-8}$ and $w_j^{critic}$ is the synaptic weights from the reservoir. The output of the critic $v(t)$ and the reward $R(t)$ in Eq. 56 define the continuous temporal difference (TD) error (Doya, 2000; Frémaux et al., 2013)

$$\delta^{DA}(t) = R(t) + \dot{v}(t) - \frac{1}{\tau_g} v(t)\,, \tag{35}$$

and discretized according to Kumar et al. (2022)

$$\delta^{DA}(t) = R(t - \Delta t) + \big[v(t) - \big(1 + \alpha_g\big)v(t - \Delta t)\big]\,, \tag{36}$$

with $\alpha_g \equiv \Delta t / \tau_g$ and, $\tau_g = 3000\ ms$ for Figures 3, 5 and 7 that did not have an obstacle and $\tau_g = 10{,}000\ ms$ for Figures 4 and 6 that required the agents to navigate past an obstacle. Synaptic plasticity of the weights onto the critic is governed by the two-factor rule

$$\Delta w_j^{critic}(t) = \eta_{critic} \cdot r_j(t) \cdot \delta^{DA}(t)\,, \tag{37}$$

with the presynaptic reservoir firing rate $r_j(t)$ modulated by the continuous TD error (Foster et al., 2000; Sutton & Barto, 2020). A three-factor rule is also possible (Frémaux et al., 2013). Synaptic plasticity of the weights from the reservoir to the actor is governed by a three-factor rule

$$\Delta W_{kj}^{actor}(t) = \eta_{actor} \cdot\ r_j(t) \cdot \rho_k(t) \cdot \delta^{DA}(t)\,, \tag{38}$$

with the product of the presynaptic and postsynaptic activity modulated by the TD error. The learning rates $\eta_{critic}$ and $\eta_{actor}$ were optimized for each task using a grid search between $10^{-6}$ to $10^{-3}$. Refer to Table. 1 for the optimized hyperparameters.

## Learning to gate working memory

Beyond the three schemas in Figure 1, learning to gate working memory is a candidate fourth schema. We specify it algorithmically and neurally here, although we did not formalize it at the computational level. A proposal for learning a task's structure is to learn to gate working memory (Hochreiter & Schmidhuber, 1997; Lloyd et al., 2012; O'Reilly & Frank, 2006; J. X. Wang et al., 2018). Since sensory cue is given only at the start of the trial in the rodent experiment, a persistent representation of the cue is necessary to learn flavor-location associations. A bump attractor has been shown to recreate the persistent working memory dynamics in the prefrontal cortex (Parthasarathy et al., 2019; Wimmer et al., 2014). The bump attractor has $N_{bump} = 54$ neurons with firing rate given by

$$u_i^{bump}(t) = ReLU\big[x_i^{bump}(t)\big]\,, \tag{39}$$

where the membrane potential $x_i^{bump}(t)$ has dynamics

$$\tau \dot{x}_i^{bump}(t) = -x_i^{bump}(t) + \chi(t) \cdot \sum_{j=1}^{M_{cue}} W_{ij}^{inwm} u_j^{cue}(t) + \sum_{h=1}^{N_{bump}} W_{ih}^{bump} \omega\left[x_h^{bump}(t)\right] + \sqrt{\tau \sigma_{bump}^2} \xi_i(t) \, , \tag{40}$$

with $\sigma_{bump} = 0.1$ and nonlinear activation function

$$\omega(x) = \begin{cases} 0, & x < 0 \\ x^2, & 0 < x < 0.5 \\ \sqrt{2x - 0.5}, & x \geq 0.5 \, , \end{cases} \tag{41}$$

The synaptic weight ${W_{ih}}^{bump}$ is defined similarly as the lateral connectivity in the actor (Eq. 31) to connect the neurons in a ring with $w_+ = 2, w_- = -10$ and $\varphi = 300$. Since the 18 cues are encoded as a one-hot vector, ${W_{ij}}^{inwm}$ is specified such that each cue activates one unit in the ring using a synaptic weight of 1 and the ${W_{ih}}^{bump}$ activates two adjacent units so that a total of three neurons form a subpopulation to persistently maintain each cue information.

The gating mechanism $\chi(t)$ controls the information flow from the sensory cues to the bump attractor by either opening the gate to update the working memory with new information or closing the gate to maintain the information persistently maintained in working memory (Lloyd et al., 2012; O'Reilly & Frank, 2006; Todd et al., 2009). There are two gating neurons, each to update or maintain working memory respectively and have the membrane potential dynamics

$$\tau \dot{\vartheta}_i(t) = -\vartheta_i(t) + \sum_{j=1}^{N} W_{ij}^{gate} r_j(t) + \sqrt{\tau \sigma_{gate}^2} \xi_i(t) \, . \tag{42}$$

A SoftMax selection rule determines the probability of selecting a particular gating action

$$P_i^{gate}(t) = \frac{exp[\beta^{gate} \vartheta_i(t)]}{\sum_k^2 exp[\beta^{gate} \vartheta_k(t)]} \, , \tag{43}$$

with $\beta^{gate} = 2$ and, $\pi_i(t) = 1$ if gating action $i$ was chosen at time t and $\pi_i(t) = 0$ otherwise. The gating mechanism then updates or maintains working memory by

$$\chi(t) = \begin{cases} 1, & \pi_1(t) < \pi_2(t) \\ 0, & otherwise \, , \end{cases} \tag{44}$$

opening $\chi(t) = 1$ or closing $\chi(t) = 0$ information flow from the sensory cues to the bump neurons. The synaptic plasticity of the weights $W_{ij}^{gate}$ is governed by a 3-factor Temporal Difference error modulated Hebbian plasticity rule

$$\Delta W_{ij}^{gate}(t) = \eta_{gate} \cdot r_j(t) \cdot \pi_i(t) \cdot \delta^{DA}(t) \tag{45}$$

using the reservoir's presynaptic activity, gating policy as postsynaptic activity and modulated by the TD error computed by the critic with $\eta_{gate} = 10^{-4}$. Whether the working memory gating rule gradually learned using this plasticity rule generalizes to other novel conditions has not been determined prior.

All synapses that were trained using a plasticity rule were initialized to zero at the start of the simulations, although synapses initialized using a small Gaussian distribution with variance less than $10^{-3}$ replicated similar results.

## Methods for meta-reinforcement learning

Mechanistic models such as our schema-based agent offer interpretability by specifying how circuit motifs implement specific computations required for navigation and one-shot learning. However, they also rely on architectural assumptions about how different systems interact, when experimental evidence is sparse. To complement this approach, we also examined a framework based on deep meta-reinforcement learning, which can give rise to internal computations that support rapid learning after extensive training across a family of related tasks (Binz et al., 2024; Botvinick et al., 2019; Kumar, Manoogian, et al., 2025; Lin et al., 2025; J. X. Wang et al., 2018) with minimal architectural bias. However, the solutions that emerge in meta-RL are often difficult to interpret and may differ from the computation the experimenter expects (Qian et al., 2024; Schaeffer et al., 2022). Using both approaches therefore provides complementary advantages (Kumar et al., 2024): the mechanistic model affords biological grounding and may be more interpretable, while the meta-RL may provide an alternative architecture that can also perform the task.

### Meta-reinforcement learning agent architecture and training

Figure S7A shows our implementation of a meta-RL agent using the architecture of J. X. Wang et al. (2018), noting that vanilla recurrent neural networks (RNN) can also support inference through dynamics alone without synaptic updates (Kumar, Manoogian, et al., 2025). At each timestep $t$, the agent received 49 place-cell activity as a vector $\boldsymbol{u}^{pc}(t)$ a one-hot cue vector $\boldsymbol{u}^{cue}$ of size 12, concatenated into the observation:

$$\boldsymbol{u}(t) = [\boldsymbol{u}^{pc}(t), \boldsymbol{u}^{cue}(t)\,], \tag{46}$$

We also implemented a second agent with privileged access to cue-specific goal information: when the agent reached a rewarded location, the cue and the corresponding place cell activity were stored in a symbolic key–value memory. On subsequent trials, the recalled goal place cell activity vector $\boldsymbol{u}^{goal}(t)$ of size 49 was concatenated with the current place cell activity and sensory cue:

$$\boldsymbol{u}(t) = [\boldsymbol{u}^{pc}(t), \boldsymbol{u}^{cue}(t)\,, \boldsymbol{u}^{goal}(t)], \tag{47}$$

This parallels Banino et al. (2018), who provided goal grid cell activity to facilitate vector-based navigation. The observation was passed to a recurrent neural network with the dynamics:

$$h_i(t) = \tanh\left(\sum_j^P W_{ij}^{in} u_j(t) + \sum_k^N W_{ik}^{rec} h_k(t-1) + b_i^{in} + b_i^{rec}\right) \tag{48}$$

where $\boldsymbol{h}(t)$ is the RNN's hidden state with $N = 256$. The critic readout was a linear projection of the hidden state:

$$v(t) = \sum_{i}^{\mathrm{N}} w_i^{critic} h_i(t) + b, \tag{49}$$

The actor was also a linear projection of the hidden state but with a softmax function which produced a distribution over eight movement actions (up, down, left, right, and diagonals).

$$a_j(t) = softmax\left(\sum_{\mathrm{i}}^{\mathrm{N}} W_{ji}^{actor} h_i(t) + b_j\right), \tag{50}$$

Agents were trained using the Monte-Carlo Advantage Actor–Critic (A2C) algorithm (Kumar, Manoogian, et al., 2025; Lin et al., 2023, 2025; Mnih et al., 2016). The advantage estimate was

$$A(t) = G(t) - v(t) = \sum_{k=0}^{T-t} \gamma^k r(t+k) - v(t), \tag{51}$$

With reward discount $\gamma = 0.99$. The full A2C objective was:

$$\begin{aligned} L_{total} &= L_\pi + L_v + L_{ent} \\ &= \sum_{t}^{T} [-\log \pi(a(t)|u(t);\, \theta, \theta_{actor}) A(t) \\ &\quad + \beta_v A^2(t; \theta, \theta_{critic}) - \beta_{ent} H(a(t); \theta, \theta_{actor})], \end{aligned} \tag{52}$$

where $\theta, \theta_{critic}, \theta_{actor}$ denote the trainable parameters of the RNN, actor and critic respectively. We used coefficients $\beta_v = 0.5$ for the value loss and $\beta_{ent} = 0.001$ for the entropy regularization. $H(a(t))$ denotes the entropy of action distribution at each time step. Gradients were backpropagated through time (BPTT) at the end of each learning trial using the Adam optimizer with learning rate $\eta = 0.0001$. The optimal learning rate was selected from $\eta = \{10^{-5}, 2.5 \times 10^{-5}, 10^{-4}, 2.5 \times 10^{-4}, 10^{-3}, 2.5 \times 10^{-3}, 10^{-2}\}$. The signs of the policy-gradient, entropy, and value-loss terms ensure that the policy is updated to maximize expected return with action exploration while the critic is trained to minimize the value-prediction error.

## Task descriptions

### Simulation 1: Random foraging

The task was to understand how the LEARN METRIC REPRESENTATION schema uses place cell activity to learn a continuous metric representation. The agent moves within a spatially continuous two-dimensional square arena bounded by walls of length 1.6m with possible agent positions $x = (\pm 0.8\, m, \pm 0.8\, m)$. At the start of each trial, the agent's current coordinate estimation $\boldsymbol{p}(t)$ was reset to zeros while its true position drawn with equal probability from midpoints of the boundary walls. The agent moves by executing time-dependent actions $\boldsymbol{a}(t)$ from a random policy that affect its velocity according to

$$\dot{\boldsymbol{s}}(t) = \boldsymbol{a}(t)\,. \tag{53}$$

Using Euler's method of discretization with time step $\Delta t$, this results in position updates

$$\boldsymbol{s}(t + \Delta t) = \boldsymbol{s}(t) + \Delta t \cdot \boldsymbol{a}(t)\,. \tag{54}$$

If the updated position ends up outside the area, the agent moves $0.01\ m$ inwards perpendicular to the closest boundary from its last position given by $\hat{\boldsymbol{a}}(t)$. The agent explored the maze over 20 trials for 300 seconds each. Synaptic plasticity from place cells to the coordinate cells was switched on according to Eq. 9.

**Simulation 2: Associating cues to coordinates**

This task was to determine the association capabilities of the LEARN FLAVOR-LOCATION schema by training a network with three readout units to associate up to 200 one-hot vector cue inputs with 200 goal coordinates randomly drawn from a uniform distribution between [-1,1]. The task was split into two phases: A) association and B) recall.

During the association phase, randomly drawn goal coordinates concatenated with a value of one, for example [0.4, -0.2, 1], was set as the target vector $\boldsymbol{g}^*(t)$. Synaptic plasticity governed by either the LMS or the EH rule was switched on for five seconds. Thereafter plasticity was switched off to determine if the network was able to maintain the learned goal coordinates for an additional five seconds. The cue-coordinate pair was presented only for one trial with plasticity switched on. Next, during the recall phase, the reservoir's internal activity was reset by drawing each unit's membrane potential from a Gaussian distribution with zero mean and variance 0.1 and the cue was presented as input to the network. The one-shot recall error was determined by taking the mean squared error between the target $\boldsymbol{g}^*(t)$ and the recalled goal readout $\boldsymbol{g}(t)$.

To delete a cue specific association, the chosen cue was presented as input to the network and plasticity modulated by various levels of acetylcholine was switched on to induce synaptic depression. The number of neurons and nonlinearity were varied to determine their effects on influencing the network's one-shot learning and recall capabilities.

**Simulation 3: Displaced match to place (DMP) task**

Following Steele and Morris (1999), the task involves navigating to a single goal in the square maze described in simulation 1. A goal location is randomly chosen out of 49 possible reward locations distributed throughout the maze such that the centers of possible locations are $0.2\ m$ from each other or a boundary. All possible reward locations are circles with a radius of $0.03\ m$. A session constitutes of four learning trials and one non-rewarded probe trial where the goal remains in the same location. In the following session, a new goal location is chosen.

The agent is free to explore the area for a maximum duration of $T_{max}$ per trial. After the agent reaches the reward, a total reward value of $R = 5$ is disbursed at a reward rate $R(t)$ defined by

$$\dot{R}_{decay}(t) = -\frac{R_{decay}(t)}{\tau_{decay}};\ \dot{R}_{rise}(t) = -\frac{R_{rise}(t)}{\tau_{rise}}\ , \quad (55)$$

$$R(t) = \frac{R_{decay}(t) - R_{rise}(t)}{\tau_{decay} - \tau_{rise}}\ , \quad (56)$$

with $\tau_{rise} = 100\ ms$ and $\tau_{decay} = 250\ ms$, similar to Kumar et al. (2022). When the agent reaches the reward, instantaneous updates

$$R_{rise}(t) \to R_{rise}(t) + R; R_{decay}(t) \to R_{decay}(t) + R\ , \quad (57)$$

are made such that $R(t)$ integrates to $R$. To prevent infinitely long trials and to model consummatory behavior, the agent remains stationary until $R - 10^{-8}$ of the reward has been consumed, thereafter the trial ends.

**Simulation 4: Single goal with an obstacle**

The actor–critic algorithm allows an agent to navigate past an obstacle for single goals (Frémaux et al., 2013), while the NAVIGATE scheme only affords direct heading. To determine if a combination of actor–critic and schema could improve an agent's navigation capability, the task was to navigate past an obstacle to a single goal found at the center of the arena with coordinates $(0, 0)$. The goal was surrounded on three sides by an inverted U-shaped obstacle with width $0.08\ m$ and length $0.6\ m$. The agent's starting positions were constrained to either the north, east or west of the arena to prevent the agent from solving the task by direct heading. The total reward value was $R = 1$.

**Simulation 5: Multiple paired associations (MPA)**

To model Tse et al. (2007), each session comprised of six trials with each of the six possible cues given to the agent in a random sequence throughout the trial. A reward was disbursed only when the agent reached the cue specific target location. Non-rewarded probe sessions were sessions 2, 9, and 16.

After 20 sessions to learn the Original Paired Associations (OPA), agents were introduced to either the OPA, 2 New Paired Associations (2NPA) with Cues 7 and 8 replacing Cues 1 and 6 and cue–locations 2–5 remained, 6 New Paired Associations (6NPA) with new Cues 11 to 16, or New Maze (NM) with the same 6 new paired associations as 6NPA but with place fields shuffled.

To model the effect of place cell remapping when an agent is introduced to a new environment with different landmarks, each place cell's selectivity within the maze was shuffled such that they were anchored to a different location. For example, if place cells 1 and 2 had peak activation at coordinates (-0.2,0) and (0.5,0.6) respectively, these cells will have peak activation at different coordinates (0.3,-0.4) and (-0.3,-0.7) in a new environment. See Figure S2 for an example. The reward value was $R = 5$.

**Simulation 6: MPA with an obstacle**

To increase the complexity of the navigation task, an obstacle was introduced to the open-field arena. The obstacle consisted of a wall with width $0.08\ m$ and length $0.8\ m$, and two parallel walls of width $0.08\ m$ and length $0.68\ m$. The obstacle did not cover the goal locations as in the original paired association, new paired association and new maze configuration described in Tse et al. (2007). The same task parameters were used as in the MPA task although the total reward value was kept at $R = 1$.

To prevent the agents from reaching the goals by direct heading, the starting positions of the agent were constrained to Table 2 so that the goal can only be reached by navigating past the obstacle. For example, the starting position for Cue 1 is randomly chosen to either be the east or south while starting position for Cue 2 is either the west or south.

**Table 2: Possible starting positions for trials with specific FLAVOR-LOCATION pairs.**

| Starting positions | East | North | West | South |
|---|---|---|---|---|
| **Cues given during trial** | 1, 4, 5, 7, 11, 13, 14 | 3, 4, 5, 6 ,8, 15, 16 | 2, 3, 6, 8, 12, 15, 16 | 1, 2, 3, 4, 7, 11, 12, 13, 15 |

**Simulation 7: MPA with transient cue and distractor**

To fully replicate the biological experimental conditions in Tse et al. (2007), the flavor cue was given to the agent one second after the trial started for two seconds. During this cue presentation period, the place cell activity and agent's actions were silenced to simulate the rat in the starting box with no knowledge of its position in the maze. The sensory cue was then switched off by setting the sensory cue vector to zero and place cell activity and agent's actions switched on for navigation. The task relevant cue was not given to the agent thereafter. Instead, distractor Cues 17 and 18 were chosen randomly and presented three seconds after navigation has commenced at the frequency of $0.2\ Hz$. The distractor was presented for one second, either once or twice within a trial.

**Simulation 8: Meta-learning task structure**

Figure S7B is a schematic of the meta-learning framework which consisted of an inner loop of training sessions followed by an outer-loop meta-update. During the inner loop, the agent completed 20 MPA sessions containing six cue–location associations, followed by two sessions each of OPA, 6NPA (six new paired associates), and NM (place cells shuffled with 6NPA combination). The reward value was one $R = 1$. The network parameters after the MPA sessions were used for OPA, 6NPA and NM independently. These parameters were updated at the end of every learning trial using the A2C objective, whereas probe trials did not contribute to learning. Probe sessions were interleaved at fixed points (Sessions 2, 9, 16, and 22 for OPA, Session 24 for 6NPA, and Session 26 for NM) to assess generalization without further weight updates.

Two variants of the task were used in Figure S7C. In the PC condition only, the place-cell selectivity was fixed across all epochs, and only the reward locations were resampled. This required the agent to acquire a vector-based strategy for navigating to new targets while relying on the stable place cell spatial code. In the Shuff-PC condition, the place-cell selectivity was randomly shuffled at the start of each epoch, replicating hippocampal remapping in a new environment. This forced the agent to learn a separate spatial representation that generalized across place cell remapping in addition to learning a vector-based solution, a prerequisite for demonstrating one-shot learning under spatial representational change.

The outer loop comprised 200 epochs. At the beginning of each epoch, a new set of 12 reward locations was sampled from the 49 possible arena positions, with six assigned to OPA cues and six to 6NPA/NM cues. Only the network weights obtained at the end of the 20 MPA sessions were used for the outer-loop update, ensuring that learning during the NPA and NM sessions did not directly influence the meta-update. This training structure required the agent to repeatedly relearn new cue–goal mappings under either a fixed spatial representation (PC condition) or a remapped spatial representation (Shuff-PC condition), thereby allowing us to dissociate the contributions of the LEARN METRIC REPRESENTATION and NAVIGATE schemas.

## Hyperparameters

In all simulations, the non-rewarded probe trials lasted 60 seconds and did not disburse reward even if the agent reached the correct reward location. Hyperparameters such as learning rates, TD error time constants and $\beta^{control}$ were optimized using grid search to minimize the latency over the initial MPA sessions and maximize the visit ratios for OPA and 6NPA.

**Table 1. Hyperparameters optimized for each task**

| Task | Figure | $R$ | $\tau_g$ (ms) | $\eta_{critic}$ | $\eta_{actor}$ | $\Omega_{Ach}$ |
|---|---|---|---|---|---|---|
| DMP | 3 | 5 | 3,000 | $2 \times 10^{-4}$ | $5 \times 10^{-5}$ | $5 \times 10^{-4}$ |
| Single goal + Obstacle | 4 | 1 | 10,000 | $1 \times 10^{-4}$ | $1 \times 10^{-5}$ | $5 \times 10^{-5}$ |
| MPA | 5 | 5 | 3,000 | $2 \times 10^{-4}$ | $5 \times 10^{-5}$ | $5 \times 10^{-5}$ |
| MPA + Obstacle | 6 | 1 | 10,000 | $1 \times 10^{-4}$ | $5 \times 10^{-6}$ | $5 \times 10^{-5}$ |
| MPA + Distractor | 7 | 5 | 3,000 | $2 \times 10^{-4}$ | $1 \times 10^{-4}$ | $5 \times 10^{-5}$ |

## Statistical analysis

Sample sizes are the number of independent simulations per agent and condition, and are stated in the main text and figure legends (n=720 for Figures 3 and 5; n=480 for Figures 4, 6 and 7). The large number of simulations may be a reason p-values are small, so we also report effect sizes using Cohen's $d = t/\sqrt{n}$ for one-sample and paired t-tests, along with 95% confidence intervals, while giving p-values only as thresholds. Visit ratios were compared against chance performance of 16.7% for six goals or 8.3% for twelve goals.

# Results

In this section, we begin by describing three schemas: (A) LEARN METRIC REPRESENTATION, which enables self-localization; (B) LEARN FLAVOR–LOCATION, which supports rapid cue–goal association; and (C) NAVIGATE, which converts recalled goals into movement directions. For each schema, we examined its conceptual description, algorithmic realization, and neural implementation using local learning rules. Using these schemas, we constructed three classes of agents: (1) a standard actor–critic agent without schemas, serving as a canonical reinforcement learning benchmark; (2) a symbolic schema-based agent; and (3) a fully neural schema-based agent. These agents allowed us to investigate whether biologically plausible schemas can support one-shot learning in spatial navigation tasks that have been experimentally proposed to depend on schema formation.

We first evaluated all agents in the displaced match-to-place (DMP) task, which probes rapid adaptation to changing goal locations. We then examined navigation to a fixed goal in the presence of obstacles, comparing basic agents with hybrid agents that combine schemas with an actor–critic policy. The core analyses focused on the two-stage multiple paired-association (MPA) task, in which rodents exhibit one-shot learning after schema acquisition. We further tested both basic and hybrid agents on an obstacle version of the MPA task to assess robustness under constrained navigation. Finally, we evaluated performance in an MPA task with transient cues and distractors to determine whether learned working memory gating generalizes across novel cue–goal associations.

Across tasks, we found that while the actor–critic agent could gradually learn through reinforcement, only schema-based agents exhibited one-shot learning in both the DMP and MPA paradigms. Hybrid schema–actor–critic agents succeeded when obstacles prevented direct navigation, while preserving one-shot learning. Moreover, learning a working memory gating policy enabled schema agents to selectively encode task-relevant cues, ignore distractors, and generalize immediately to new paired associations. Together, these results reproduce key behavioral signatures of schema-dependent learning observed experimentally (Tse et al., 2007) and are consistent with reported neural plasticity dynamics following schema acquisition (Tse et al., 2011).

**A) LEARN METRIC REPRESENTATION SCHEMA: Learn a metric representation of the environment to self-localize and for NAVIGATE**

| Computation | Neural implementation |
| --- | --- |
| Use self-motion information and place cell activity to learn a continuous X–Y coordinates–based metric representation for vector–based navigation. | Place cells are used as input to the X and Y coordinate cells (Eq. 4) as outputs. The synapses from place cells to coordinate cells are gradually learned using a path integration temporal difference error (Eq. 7) modulated learning rule (Eq. 9) with eligibility trace (Eq. 8). |

**Algorithm**
Estimate start coordinates
**For** t < T:
  Move in direction specified by policy
  Estimate coordinates using place cells
  Compute path integration error
  Compile history of place cell activity
  Minimise path integration error
**End**

$$\tau\dot{\mathbf{p}}(t) = -\mathbf{p}(t) + W^{MR}\cdot\mathbf{u}^{pc}(t) \quad (4)$$

$$\delta^{MR}(t) = \mathbf{p}(t) - \mathbf{p}(t-\Delta t) - \hat{\mathbf{a}}(t) \quad (7)$$

$$\tau_{MR}\dot{\mathbf{e}}(t) = -\mathbf{e}(t) + \mathbf{u}^{pc}(t) \quad (8)$$

$$\Delta W^{MR}(t) \propto \mathbf{e}(t)\delta^{MR}(t) \quad (9)$$

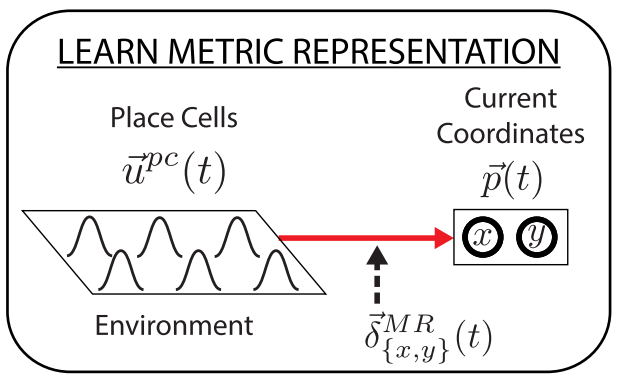

**B) LEARN FLAVOR–LOCATION SCHEMA: Associate flavor cue to reward location after one trial to recall goal for NAVIGATE**

| Computation | Neural implementation |
| --- | --- |
| Associate the flavor cue with the coordinates at which a reward was disbursed after a single trial. Use flavor cues to recall the goal coordinates. | The cue vector is passed as input to a hidden layer with three output units representing X, Y goal coordinates and reward recall value (Eq. 17). Only the output synapses are subject to the reward modulated Exploratory Hebbian rule (Eq. 20). Associations are depressed using acetylcholine modulated Hebbian learning (Eq. 25). |

**Algorithm**
Initialize associative memory and get cue
**For** t < T:
  Use cue to query the keys to recall associated goal coordinates and reward value from the value matrix
    **If** no reward disbursed:
      Delete flavor–location association
  **Else**
    Explore maze using default policy
    **If** a reward is disbursed:
      Store cue in key matrix, and agent's estimated current coordinates and reward in value matrix
**End**

$$\tau\dot{\mathbf{g}}(t) = -\mathbf{g}(t) + W^{FL}\cdot\mathbf{r}(t) + \xi^{FL}(t) \quad (17)$$

$$\mathbf{g}^{*}(t) = [\mathbf{p}(t), \Theta[R(t)]] \quad (18)$$

$$\Delta W^{FL}(t) \propto \mathbf{r}(t)\cdot[\mathbf{g}^{noisy} - \mathbf{g}(t)]\cdot M(t)\cdot\Theta[R(t)] \quad (20)$$

$$M(t)\begin{cases}1, \bar{P}(t) < P(t)\\ 0, otherwise\end{cases} \quad (21)$$

$$P(t) = -\sum[\mathbf{g}^{*}(t) - \mathbf{g}^{noisy}(t)]^{2} \quad (22)$$

$$\Delta W^{FL}(t) \propto \mathbf{r}(t)\cdot\mathbf{g}(t)\cdot -\Omega_{Ach} \quad (25)$$

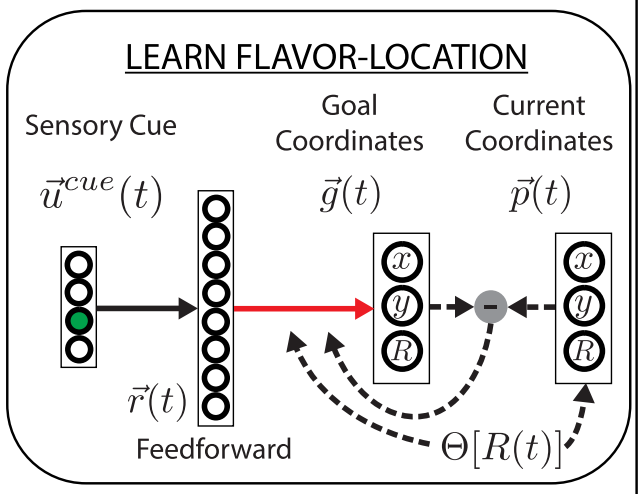

**C) NAVIGATE SCHEMA: Move from arbitrary location to goal location**

| Computation | Neural implementation |
| --- | --- |
| Take in the agent's current coordinates, recalled goal coordinates and reward recall value as inputs to output direction to move. | Neural network with two nonlinear hidden layers pre-trained by backpropagation on a dataset with different current coordinate, goal coordinate, reward recall values and action. Weights were fixed during task learning. |

**Algorithm**
Get agent's current coordinates
Get goal coordinates and transformed reward
**For** t < T:
  **If** recalled reward > threshold (Eq. 38):
    Compute vector subtraction between current and goal coordinates (Eq. 36)
    Choose action based on direction specified by vector (Eq. 37)
**Else**
  No action selected
**End**

$$\mathbf{d}_{\{x,y\}} = \mathbf{g}_{\{x,y\}} - \mathbf{p}(t) \quad (26)$$

$$\mathbf{q}^{NAV}(t) = \mathrm{softmax}(K^{actions}\cdot\mathbf{d})\cdot\epsilon(t) \quad (27)$$

$$\epsilon(t) = \begin{cases}1, 0.6 < g_{j=3}(t)\\ 0, otherwise\end{cases} \quad (28)$$

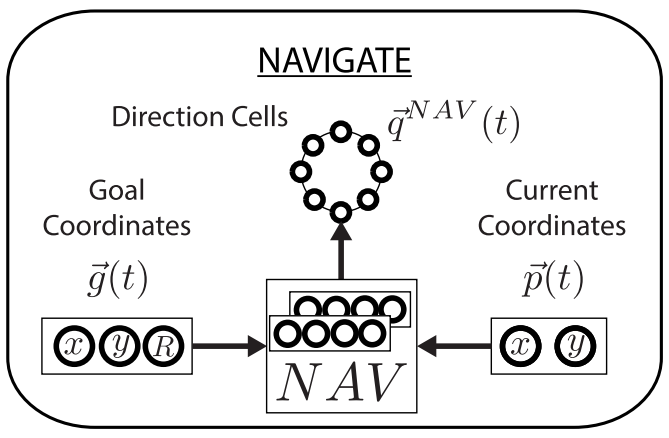

**Figure 1. Schemas for learning one-shot navigation to multiple goals.** A schema is a framework that describes relationships between information and actions. Each row summarizes a schema conceptualized as a computation in Marr's sense, with a corresponding algorithm (left), key equations (center), and connectivity schematic (right) of a biologically plausible neural implementation.

## Schemas for one-shot navigation to goals

To formalize how prior knowledge enables rapid learning, we adopted Marr's three-level framework for understanding neural computation: (1) conceptually, as the computational goal specifying the input and desired output or goal, (2) the algorithm as a pseudocode that carries out the computation, and (3) the physical implementation of biological mechanisms that exactly or approximately realize the computation and algorithm (Marr, 2010; Marr & Poggio, 1977; Poggio, 2012). We propose three schemas that together support one-shot navigation to goals. For each schema, we describe its computational role, outline an algorithmic solution, and demonstrate a biologically plausible neural implementation (Fig. 1).

### LEARN METRIC REPRESENTATION schema

The computation performed by the LEARN METRIC REPRESENTATION schema is to acquire a continuous, low-dimensional metric representation of the agent's position in a novel environment (Fig. 1A). This enables an agent to estimate its coordinates for self-localization which can then be used for vector-based navigation without needing a lookup table. This additional mechanism is needed because although place cell or grid cells represent an agent's current position, these representations alone do not directly support computing a vector from the current location to an arbitrary goal (Bush et al., 2015; Fiete et al., 2008).

Algorithmically, the schema learns to map binned place cell activity to continuous Cartesian coordinates by minimizing discrepancies between predicted and actual displacement (refer to Fig. 1A for the pseudocode). We implemented a biologically plausible neural model adapted from Foster et al. (2000), in which place cells project to X- and Y-coordinate neurons (Eq. 4). A path-integration–derived temporal-difference (TD) error is computed by comparing estimated displacement between current and previous position with self-motion corrected for boundary conditions (Eq. 7). Synaptic plasticity depends on an eligibility trace of presynaptic activity (Eq. 8) modulated by this TD error (Eq. 9).

As the agent randomly explored the arena, the path integration-derived TD error gradually decreased (Figure S1A), and synaptic weights converged to a stable solution. The learned coordinate neurons exhibited monotonic tuning along the horizontal and vertical axes (Fig. 2A, top), resulting in increasingly accurate position estimates along arbitrary trajectories (Fig. 2A, bottom). This learned metric representation was subsequently used by the NAVIGATE schema to compute goal-directed movement. When the agent was transferred to a new environment where place cell selectivity is remapped, the LEARN METRIC REPRESENTATION schema continues to learn a stable metric representation (Figure S2) for one-shot vector-navigation (Figure S4).

### LEARN FLAVOR–LOCATION schema

The LEARN FLAVOR–LOCATION schema performs the computation of associating a sensory cue with the corresponding goal location after a single rewarded trial (Fig. 1B). When reward is received, the agent's current estimated coordinates define the goal associated with the cue. The same cue presentation in the subsequent trial should recall and maintain the associated goal coordinates with high fidelity.

At the algorithmic level, this computation can be implemented using a key–value memory. After a successful trial, the cue is stored as a key, and the corresponding goal coordinates and a recall value of one are stored as values (Fig. 2B). On subsequent trials, the cue serves as a query, and similarity-based retrieval (Eq. 14) determines the recalled goal (Eq. 15). If recall is accurate, the recall value will be close to 1; if recall is inaccurate, the recall value will be closer to 0. If navigation based on the recalled goal fails to yield reward, the association is selectively deleted with recall value close to 0, allowing the agent to return to exploration without disrupting other stored associations.

We implemented this schema using a biologically plausible hetero-associative memory network consisting of either a nonlinear feedforward layer or a recurrent reservoir projecting to three readout neurons encoding the goal's X-coordinate, Y-coordinate, and recall value (Eq. 17). One-hot encoded cue inputs were projected through fixed random synaptic weights to gate network activity into separable cue-specific subspaces, thereby reducing interference between cue-coordinate associations for continual one-shot learning (Fig. S1C). Plasticity of readout synapses was governed either by a reward-modulated least mean squares (LMS) delta rule (Eq. 19) (Hoerzer et al., 2012; Sussillo & Abbott, 2009) or by a more biologically plausible 4-factor reward-modulated exploratory Hebbian (EH) rule adapted from Hoerzer et al. (2012). The EH rule combines presynaptic activity, postsynaptic activity, a binary variable error signal $M(t)$ indicating whether outputs approach a target vector (concatenation of the agent's current coordinates and recall value) (Eq. 18), and a binary variable $\Theta[R(t)]$ indicating reward presence. Reward gating by $\Theta[R(t)]$ ensures that associations are formed only at rewarded locations.

Figure 2C shows feedforward networks with a tanh nonlinearity and with three goal neurons (X, Y, Recall) trained to learn 200 cue-coordinate associations sequentially for one trial each. Plasticity was switched on to learn an association. Thereafter, plasticity was switched off, and the same cue was given as input to test the recall mean squared error (MSE) of the goal coordinates and the recall value. A higher area under the curve (AUC) corresponds to poorer recall MSE performance. Learning curves are shown for readout synapses trained using either the powerful local mean squared (LMS) algorithm or the more biologically plausible exploratory Hebbian (EH) rule, with learning rates tuned separately for each architecture and learning rule to minimize recall MSE AUC (See Figure S3).

Increasing network size significantly reduced recall error for both learning rules, with a gradual increase in recall MSE as the number of cue-coordinate associations increased. While LMS achieved lower recall AUC than EH at smaller network sizes, the more biologically plausible EH rule matched LMS performance at larger sizes, with statistically significant results across seeds. These patterns held across feedforward and reservoir architectures and nonlinearities (tanh, $\phi^A$, and ReLU; Figure S3) when learning rates were optimally tuned. Still, a reservoir network with $N = 1024$ and $\phi^A$ nonlinearity demonstrated the lowest recall AUC. Hence, the association capacity depends on network architecture, number of neurons, and learning rule. Importantly, these networks showed a monotonic increase in recall MSE indicating smooth association capacity–accuracy tradeoff without a sharp memory cliff, consistent with recent hetero-associative networks (Sharma et al., 2022; Tyulmankov et al., 2021).

It is also possible to delete a specific cue-coordinate association, as the key-value matrix algorithm could, with biologically plausible synaptic plasticity. To do so, we introduce synaptic depression governed by a previously proposed 3-factor acetylcholine-modulated Hebbian rule (Eq. 25) (Zannone et al., 2018). Figure 2D shows that when the associative EH rule was switched on (red dashed line) for a reservoir to learn cue-coordinate association, the readout units converged to specific X (blue trace), Y (orange trace) coordinates and a recall value of 1 (green trace) within 4 seconds, which were maintained after plasticity was switched off (black dashed line). When the reservoir was reset with random activity and the same cue was presented, the readout units recalled the associated goal coordinates and the recall value was close to 1.

When different amounts of acetylcholine were introduced ($Ach = 0.1, 0.01, 0.001$) for five seconds (magenta to black dashed lines) for Cues 1, 2 and 3, the activity of the three readout units converged towards zero at different rates, representing deletion of cue–coordinate association. Specifically, acetylcholine at levels 0.1, 0.01 and 0.001 need to be introduced for 1, 7, and 45 seconds so that there will be sufficient time for synaptic depression to cause the recall value (green trace) to fall below the 0.6 threshold respectively. Importantly, deleting cue specific associations did not affect the recall of other cue–coordinate associations (Fig. 2D, Cue 4) demonstrating the network's flexibility in writing and deleting paired associations without disrupting other associations using biologically grounded learning rules.

Consistent with experimental and theoretical work showing elevated tonic acetylcholine during exploratory behavior (Brzosko et al., 2017; Zannone et al., 2018), acetylcholine-modulated Hebbian plasticity is persistently active during exploration in the LEARN FLAVOR–LOCATION schema (Table 1). Reward delivery transiently engages reward-modulated exploratory Hebbian learning to write new associations, whereas tonic acetylcholine selectively weakens existing associations, allowing the agent to discard incorrect cue–goal associations and return to exploration without disrupting intact memories.

### NAVIGATE schema

NAVIGATE is a schema whose computation allows the agent to head directly to a recalled goal coordinate from any location, even if it had not traversed that path. It can be performed by a 3 step algorithm: (1) vector subtraction (Eq. 26) between the goal and current coordinates to determine the distance and direction to the goal from the current location, (2) choose the relevant action (Eq. 27) out of all possible actions that will bring the agent closer to the goal via direct heading, and (3) suppress the chosen action (Eq. 28) if the recall value $g_{i=3}(t)$ falls below a threshold value of 0.6 (Fig. 1C). In other words, vector subtraction and the corresponding action is enacted only if the recalled value is greater than a threshold value of 0.6. If the recalled value falls below the threshold, the output vector is suppressed by returning a zero vector, without explicitly specifying the direction of movement. This implements a simple uncertainty-gating mechanism: when the recalled value is weak, the model withholds directional action rather than committing to a potentially incorrect goal vector.

To obtain a neural implementation of NAVIGATE, we pretrained a network with two nonlinear hidden layers, each with 128 units on a dataset comprising different current coordinates, goal coordinates and recall values as inputs and the relevant action to take, computed symbolically (Eq. 26–28) as outputs (refer to Methods). The network weights were fixed throughout the DMP and MPA tasks (purple arrows in Fig. 1C and Fig. 3A). This is a reasonable schema as we do not think animals learn vector subtraction for goal-directed navigation only during the learning of the MPA task, and instead it could have been instantiated through development or evolution. Instead of pre-training the NAVIGATE schema, it might also be possible to hand construct this neural circuitry (Dan et al., 2024).

To characterize the learned representation, we examined neural activity in the penultimate layer across combinations of current and goal locations (Fig. 2E; Figure S6). Distinct populations of neurons exhibited tuning resembling cue-invariant distance cells, whose activity increased monotonically as the agent approached any goal, and cue-invariant angle cells with Gaussian directional tuning independent of goal location. These response profiles mirror vector-based navigation signals observed experimentally (Sarel et al., 2017), indicating that the NAVIGATE schema captures key features of biologically observed goal-direction coding.

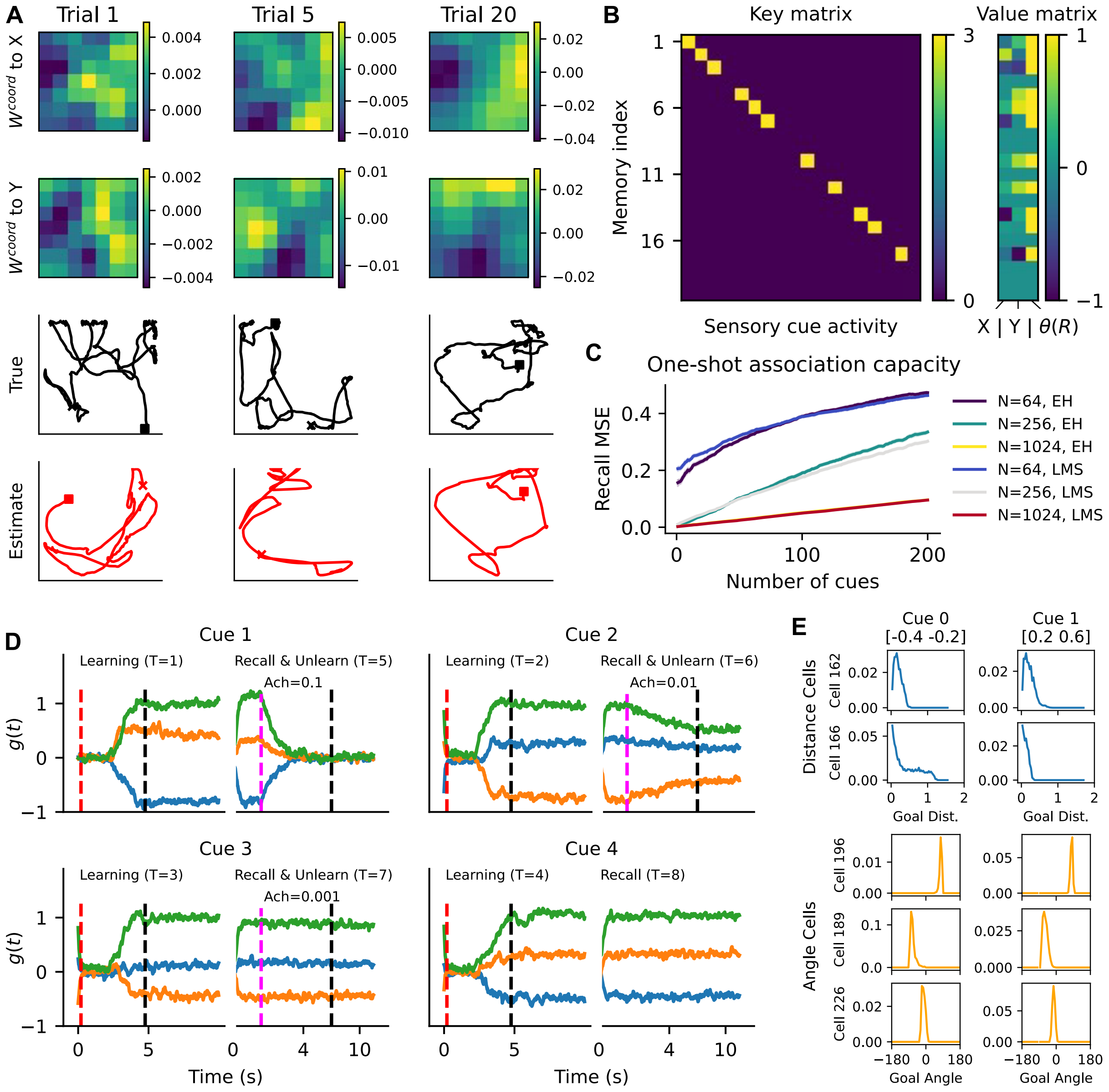

**Figure 2. Representations learned using LEARN METRIC REPRESENTATION or LEARN FLAVOR-LOCATION schemas. (A)** As an agent moves randomly in a square arena, the synaptic weights from 49 place cells to the X and Y coordinate cells converge to a stable metric representation demonstrating a monotonic increase in the firing rate of the X or Y coordinate cells respectively (top). Estimated coordinates become qualitatively similar to the true coordinates with learning (bottom). **(B)** The symbolic LEARN FLAVOR-LOCATION algorithm uses a key-value matrix to store the cue vector in each row of the key matrix and the value matrix stores the X, Y goal coordinates (between –1 to 1) and recall value (between 0 to 1). A distance metric is used to recall cue associated goal coordinates. **(C)** To study the one-shot association capacity of LEARN FLAVOR-LOCATION, plasticity was switched on during association and switched off during recall. The network could store and recall up to 200 cue–coordinate associations after a single trial, while showing smooth capacity–accuracy tradeoff. Increasing the network size decreased the recall mean squared error (MSE). The EH rule achieved similar recall MSE as the more powerful but biologically implausible LMS rule. There are 30 simulations per plot. Shaded area indicate standard error. **(D)** Using a reservoir network and EH plasticity, four cue–coordinate associations were sequentially stored across 4

trials followed by sequential recall and deletion of each cue in 4 more trials. Each panel shows the activity of goal neurons (X coordinate – blue, Y coordinate – orange, recall value – green) corresponding to association, recall and deletion of a particular cue. During association, the EH rule was switched on for 5 seconds (red dashed line to black dashed line) to associate a cue vector with a randomly sampled set of coordinates. Thereafter plasticity was switched off and the goal neurons maintained the activity. During recall, the cue input recalled the associated coordinate. Acetylcholine-modulated Hebbian learning rule was switched on (magenta dashed line to black dashed line) to induce synaptic depression for a specific paired associate. Higher acetylcholine values increased the rate of synaptic depression. The sequential deletion for Cues 1 to 3 did not affect accurate recall of Cue 4 coordinates. **(E)** Example neurons in NAVIGATE schema demonstrating cue-invariant tuning characteristics of distance and angle cells anchored to different goals [-0.4, -0.2] and [0.2, 0.6]. See Figure S6 for more cues and cells.

## One-shot learning of a single displaced goal

We begin by testing whether agents composed from symbolic or neural schemas exhibit one-shot learning in a delayed match-to-place (DMP) task, and compare their performance to a non-schema-based actor–critic agent trained by reinforcement learning (Kumar et al., 2022). In this task, agents navigated to a single reward location that was displaced every five trials; the fifth trial of each session was an unrewarded probe lasting 60 seconds. Across all trials and sessions, agents received the same sensory cue (e.g. Sensory Cue 1), requiring them to unlearn the previous goal location and rapidly learn the new one. This setting is more demanding than learning distinct cue–goal associations (Kumar et al., 2022), as it explicitly tests rapid relearning.

Figure 3A schematically illustrates the architectures of the actor–critic, symbolic schema, and neural schema agents. All agents received place-cell inputs encoding position, and the sensory cue. They have a ring-attractor based actor module that controlled movement direction and speed. The actor–critic agent (Fig. 3A, left) additionally learned a value function and updated its policy incrementally via TD error-modulated Hebbian plasticity (Eq. 36). Contrastively, the symbolic and neural schema agents learned an explicit metric representation of space, and associated the sensory cue with goal coordinates using either a Key-Value matrix or a hetero-associative network using the reward-modulated EH rule respectively. These agents used a symbolic or neural implementation of vector subtraction to output the distance and direction of movement.

Figure 3B shows that the actor–critic agent gradually learned to navigate to the single goal in the first session (Trials 1 to 4), but the latency increased after every goal displacement and plateaued within each session. Importantly, the agent did not show a significant decrease in latency between the first and second trials for any sessions, failing to demonstrate one-shot learning. In contrast, the symbolic and neural schema agents demonstrated clear evidence of one-shot learning. After early training sessions, latency reliably decreased between the first and second trials that became increasingly apparent after Session 4 (Trial 16 in Fig. 3B left). The average latency savings between Trials 1 and 2 during Sessions 5 to 9 was near zero for the actor–critic agent ($-2.5 \pm 8\ s$), but large and comparable between the symbolic ($56 \pm 9\ s$) and neural ($57 \pm 8\ s$) schema agents (right plot in Fig. 3B). These results demonstrate that schema-based agents rapidly relearned the displaced goal after a single rewarded trial.

Probe trials further revealed qualitative differences in navigation strategies (Fig. 3C). The actor–critic agent continued to navigate to the prior learned goal location instead of the new goal locations despite target displacement. This is because the synaptic weights are incrementally updated to converge to a particular value and policy map, and becomes less adaptable with prolonged training (Dohare et al., 2024). The symbolic and the neural agents initially show indirect trajectories, but directness of navigation to the goal improves, becoming apparent from around PT4.

Without acetylcholine modulated Hebbian plasticity to induce synaptic depression, schema agents perseverate to navigating to the same erroneous location although there is no reward. Hence, the acetylcholine modulated plasticity is necessary to reduce perseveration. The combination of reward and acetylcholine modulation is similar to the positive and negative RPE computed by the model-free critic as a biologically plausible forgetting mechanism.

The transition from gradual to one-shot learning reflects the separation of learning timescales across schemas. Early in training, performance improved gradually as agents acquired a stable metric representation, while cue–goal associations could be updated rapidly. Once LEARN METRIC REPRESENTATION converged, adaptation to goal displacement depended primarily on updating LEARN FLAVOR-LOCATION, enabling relearning after a single experience.

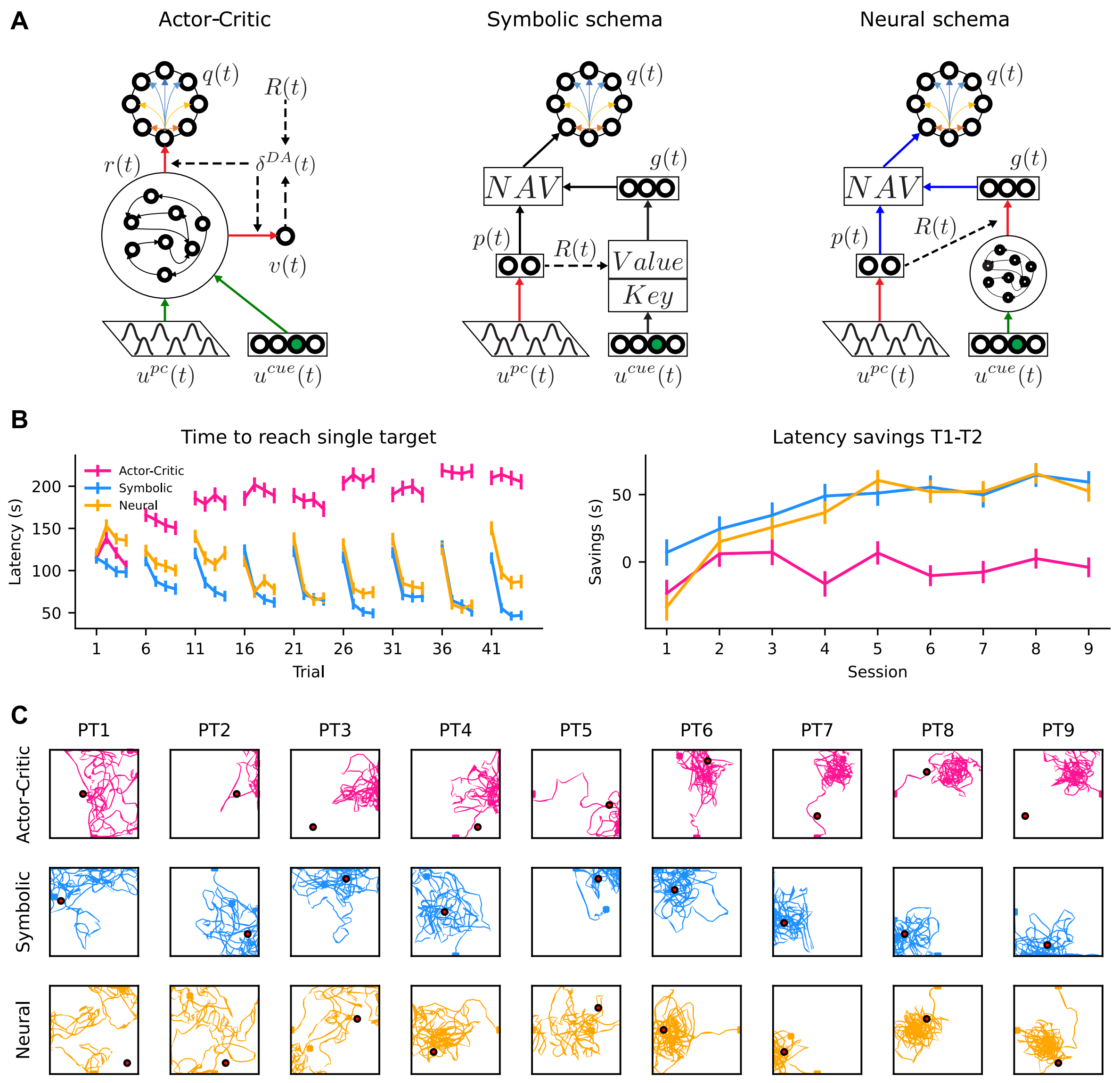

**Figure 3. One-shot learning of delayed match to place (DMP) task by schema agents. (A)** Schematic architectures of actor–critic (left), symbolic schema (center), and neural schema (right) agents. Synapses from the reservoir to the actor and critic were learned using forms of TD error-modulated plasticity. The red, green and dark blue arrows indicate synaptic weights updated using biologically plausible learning rules, drawn from a random uniform distribution and pretrained using backpropagation respectively. In both symbolic and neural agents, synapses from place cells to coordinate cells were learned using path integration TD error-modulated Hebbian plasticity. The symbolic agent used a symbolic key-value associative memory system. The neural agent used a reservoir with synapses to goal neurons trained using reward-modulated EH plasticity to store and recall goal coordinates. For the DMP task, all agents received the same Cue 1 throughout the trial for all trials, and navigated to a single target that is displaced every 5 trials, with the last trial being an unrewarded probe. **(B)** Latency (left) and savings between Trials 1 and 2 (right) to reach target. The actor–critic agent initially learns to navigate to single targets, but fails to exhibit one-shot adaptation after displacement. Both symbolic and neural schema agents demonstrate robust one-shot learning that becomes increasingly apparent after session 4. There are 720 simulations per agent. Error bars indicate 95% confidence intervals. **(C)** Example agent trajectories during unrewarded probe trials

conducted after every four training trials. Despite goal displacement, the actor–critic agent perseverates at previously rewarded locations, whereas symbolic and neural schema agents increasingly navigate toward the displaced target locations, demonstrating rapid adaptation.

## Faster navigation past an obstacle to single goal by hybrid agents

Having shown that symbolic and neural schema agents exhibit one-shot learning for displaced goals, we next examined their ability to navigate to a single goal in the presence of an obstacle. The goal and obstacle locations are shown in Fig. 4A and Fig. 4C. To prevent a trivial solution, agents were initialized from the north, east, or west boundaries of the arena, such that a direct path to the goal was obstructed. Training consisted of 60 trials interleaved with unrewarded probe trials. In addition to the pure actor–critic, symbolic, and neural schema agents (Fig. 3A), we introduced two hybrid agents. The actor–critic–symbolic agent used the symbolic schema implementation (Fig. 4A, left) while the actor–critic–neural agent used the fully neural schema implementation (Fig. 4A, right). In these hybrid agents, action selection was jointly influenced by the actor–critic and the NAVIGATE schema parameterized by $\beta^{control}$, allowing a mixture of learned policy and goal-directed control (Refer to Methods).

Behavioral performance is summarized in Fig. 4B. The actor–critic, actor–critic–symbolic, and actor–critic–neural agents showed robust learning, with navigation latency decreasing over training and the time spent at the goal increasing across probe trials (n=480 simulations per agent, t-tests for all 3 agents, $p<10^{-4}$). In contrast, pure schema agents performed poorly in this task. The symbolic agent's latency increased, while the neural schema agent showed only modest improvement and spent less time at the goal during probe trials. Thus, while pure schema agents excelled at direct goal navigation, they struggled to adapt their behavior in the presence of an obstacle.

Figure 4C shows the example trajectories (top) and value and policy maps for all agents during PT3. The actor–critic navigated past the obstacle as it learned actions specific to the location in the arena. Contrastively, the symbolic agent initially reached the goal, but performance worsened after PT1. During the initial trials, the schema agent was still learning the metric representation, hence the direct heading specified by NAVIGATE in PT1 was suboptimal and close to random (Fig. 4C, blue trajectory). As the metric representation converged, NAVIGATE specified direct heading to the goal but became stuck at the obstacle (Fig. 4C, green and red trajectories), and the goal coordinate was deleted at the end of the trial when no reward was disbursed. The neural agent demonstrated more variable trajectories as stochasticity and gradually decaying recall value below the NAVIGATE threshold suppressed direct heading (Figure S1B, 5). Nonetheless, the neural agent performed more poorly than agents with an actor–critic component.

Hybrid actor–critic–schema agents combined the advantages of both approaches. Their trajectories, value and policy maps reflected a mixed strategy in which actor–critic–derived policies guided navigation around the obstacle, while schema-based goal representations enabled direct heading once the obstacle was cleared. As a result, hybrid agents reached the goal more reliably and spent more time at the goal during probe trials than the pure actor–critic agent, demonstrating that combining learned policies with explicit goal-directed control improves navigation in structured environments.

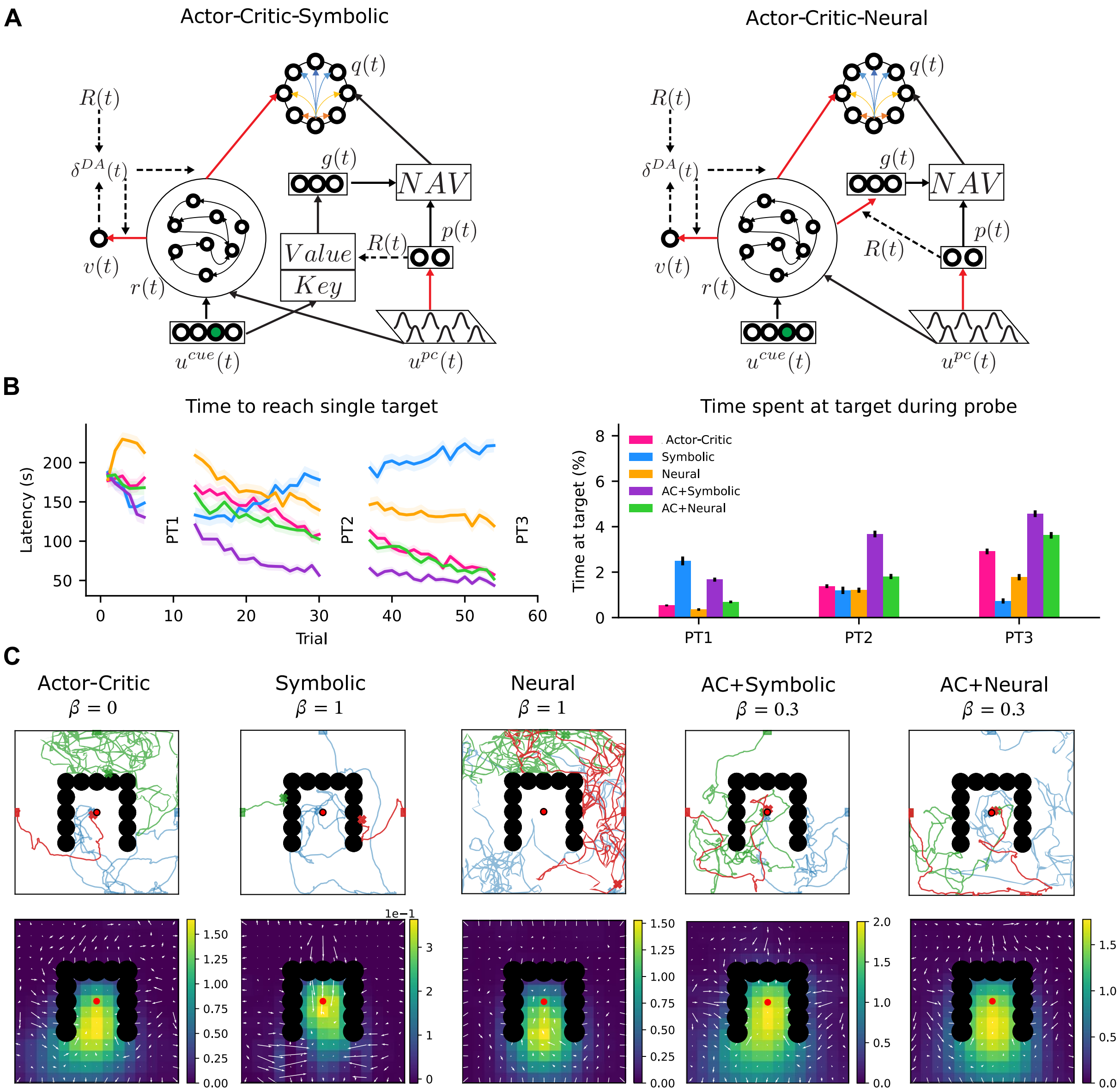

**Figure 4. Navigating past an obstacle to a single goal. (A)** Hybrid actor–critic-schema architectures combining an actor–critic with either symbolic (left) or neural (right) schemas. **(B)** Navigation latency (left) and proportion of time spent at the goal during probe trials (right). The actor–critic (pink), actor–critic–symbolic (purple), and actor–critic–neural (green) agents show decreasing latency and increased goal occupancy, whereas pure symbolic (blue) and neural (orange) schema agents struggle to navigate past the obstacle. **(C)** Example trajectories (top) during Probe Trials 1 (blue), 2 (green), and 3 (red) show both actor–critic ($\beta = 0$) and actor–critic-schema agents ($\beta = 0.3$) chose actions to navigate past the obstacle and towards the goal, while the pure schema agents ($\beta = 1$) moved only by direct heading and got stuck at the obstacle. The policy (white arrows) and value (colormap) maps (bottom), respectively, show actor and critic outputs. Hybrid actor–critic schema agents learned mixed policies that had aspects of both actor–critic and schema-based control. 480 simulations were performed for each type of agent, with the shaded area and error bars indicating 95% confidence intervals.

## One-shot learning of multiple paired association navigation

Having established one-shot learning for a single displaced goal, we next tested whether agents could rapidly acquire multiple flavor–location paired associations (MPA). The task comprised two stages. Stage 1 requires learning six flavor-location associations over 20 sessions. Each session consisted of six trials across which the agent was exposed to sensory cues 1–6 in random order. Each cue was presented throughout the trial, and the agent received a reward only if it reached the corresponding goal (insets in Fig. 5A and Fig. 5B show reward locations). Learning was interspersed with three non-rewarded probe sessions during Session 2 (PS1), Session 9 (PS2), and Session 16 (PS3). Stage 2 requires learning two or six new paired associations in a familiar or novel environment for one session. These were the six original paired associations (OPA), two new paired associations (2NPA), six new paired associations (6NPA), or a new maze (NM) condition with place fields remapped. A non-rewarded probe session followed for each condition. (See Methods)

Figure 5A shows that in Stage 1, all agents gradually learned all six flavor-location pairs. By Session 20, the latency required to reach all six goals gradually decreased to $25 \pm 1.6\ s$ (Actor–critic), $46 \pm 2.6\ s$ (Symbolic), and $67 \pm 2.9\ s$ (Neural). Visit ratio was calculated as the amount of time spent within $0.1\ m$ of the correct goal divided by the time spent within $0.1\ m$ of any of the six possible goals during a probe session. A visit ratio of $16.7\%$ was consistent with chance performance, where the agent visited all six reward locations equally or visited one location regardless of cue. All agents showed monotonic improvements in visit ratios from PS1 to PS3 (Fig. 5B, right) indicating successful gradual learning of the six paired associations.

Figure 5B shows Stage 2 which dissociated gradual learning from one-shot learning. All agents performed well in the OPA condition in which one-shot learning was not required, as the paired associates were the same as in the first stage. Importantly, only the symbolic and neural schema agents demonstrated one-shot learning of two or six new paired associations, with above chance visit ratios for 2NPA (symbolic: d=4.0, neural: d=2.6) and 6NPA (symbolic: d=2.5, neural: d=5.2) conditions (n=720 simulations per agent, $p<10^{-4}$). The actor–critic agent did not show one-shot learning, with chance visit ratios for both 2NPA (d=-0.71) and 6NPA (d=-0.22) conditions. Example trajectories for the agents during the first stage and the various second stage conditions are shown in Figure 5D.

Value and policy maps revealed the source of this behavioral difference. All agents had learned cue-specific policy and value maps for each of the six paired associates in the OPA condition (example maps shown for Cue 2 in Fig. 5E). A peak in the value map corresponds to the cue-specific location and the policy map demonstrates the path to reach the correct target. However, only the schema agents demonstrated cue-specific value and policy maps for each of the new paired associates after one session in the 2NPA or 6NPA conditions. The actor–critic maps for new cues were not sufficiently distinct, contributing to the chance performance (example maps for Cue 7 and Cue 16 shown in Fig. 5E).

In the NM condition, all agents performed poorly after one session with chance-level visit ratios (Fig. 5B). For the schema agents, shuffling place-cell selectivity invalidated the metric representation learned in the first stage, preventing accurate coordinate estimation (Figure S2) and abolishing one-shot learning in the second stage. The actor–critic resorted to explorative trajectories (Fig. 5D). When training in NM was extended by 20 sessions as Stage 3, all agents gradually relearned the six paired associations with Cues 11–16 (Figure S4). Importantly, when six additional paired associations were introduced for a single session as Stage 4 (6NPANM with Cues 7–10, 17, and 18), schema agents, but not actor–critic, again demonstrated one-shot learning confirming continual acquisition of new metric representations in new environments.

To determine the validity of the proposed schemas, we trained actor–critic agents with and without the FLAVOR–LOCATION association matrix using a meta-reinforcement learning framework (Refer to Figure S7). Meta-RL agents were repeatedly required to learn six flavor–location paired associations followed by test conditions corresponding to OPA, 6NPA and NM conditions. While the sensory cue identities were held fixed, reward locations were resampled across environments, requiring agents to generalize across spatial configurations. All agents gradually learned the original six paired associations. However, only agents endowed with associative memory exhibited reliable one-shot learning in the 6NPA condition. In these agents, partial cue-invariant distance and angle representations emerged (Figure S8), resembling the computation implemented by the NAVIGATE schema (Fig. 2E). Importantly, when place-cell selectivity was shuffled across environments, even agents endowed with associative-memory failed to demonstrate one-shot learning. This result indicates that the meta-RL framework alone might be insufficient to continually learn a new metric representation per environment (Banino et al., 2018; Schaeffer et al., 2022; Whittington et al., 2020), unlike our proposed LEARN METRIC REPRESENTATION schema.

To further study the schema agents' one-shot learning capacity, 12 new paired associates were introduced for a single trial each after Stage 1 (Fig. 5C). For each agent simulation, the goal locations for the 12 paired associates were randomly chosen out of the 43 remaining reward locations, after excluding the six reward locations that were used for the OPA condition. We arbitrarily defined a paired associate to have been learned if an agent achieved a visit ratio of more than 16.7% for the pair, well above the 8.3% expected if all 12 goals were visited randomly (see Figure S4D for example trajectories for 12NPA). After just one session of learning 12PA, the actor–critic only learned $1.9 \pm 0.17$ paired associates, whereas the symbolic schema agents learned $11.6 \pm 0.12$ paired associates.

Neural schema agents either with a feedforward layer or reservoir trained using the LMS or EH rules demonstrated a monotonic increase in one-shot learning performance when the number of hidden or recurrent units were increased from 128 to 2048. For example, the agent with a reservoir of 128 units trained with the EH rule learned $3.5 \pm 0.11$ paired associates after one trial but the one-shot learning capacity increased to $7.8 \pm 0.16$ with 1024 units. Interestingly, agents with a reservoir outperformed agents with a feedforward layer. Although the LMS rule affords better one-shot learning performance, the EH rule is still effective. At larger reservoir sizes (2048 units), capacity depended on learning rate optimization where the one-shot learning capacity decreased to $3.9 \pm 0.17$, but increased to $8.7 \pm 0.18$ when the learning rate $\eta_{goal}$ was reduced from $7.5 \times 10^{-5}$ to $5 \times 10^{-5}$.

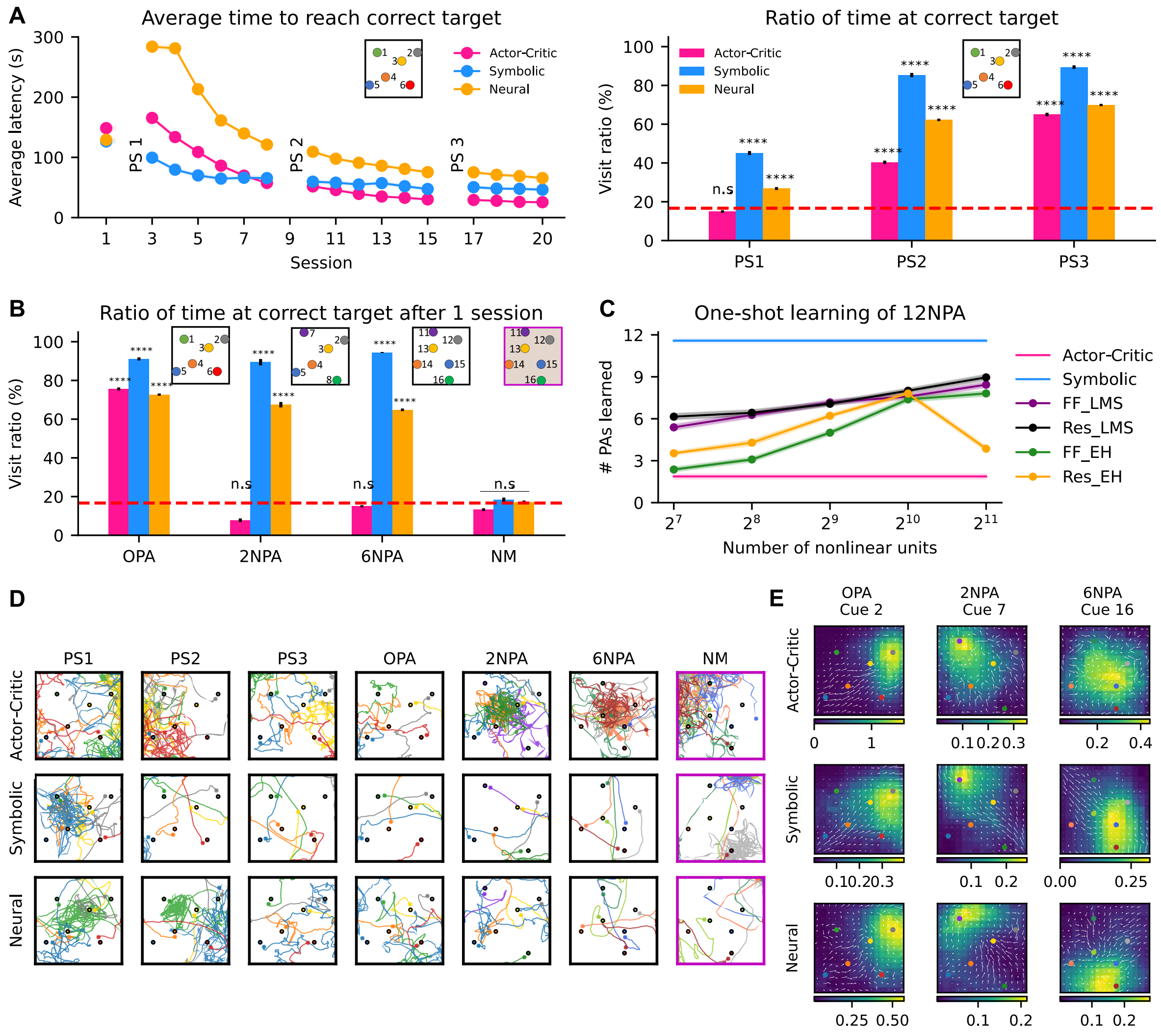

**Figure 5. Gradual then one-shot learning of multiple new paired associations.** The task was trained in 2 stages. Stage 1 consisted of 20 sessions, each with six trials in which the agent was exposed to six cues. Agents received a reward only if they reached the corresponding goal. Training was interspersed with three non-rewarded probe sessions during Session 2 (PS1), Session 9 (PS2), and Session 16 (PS3). Stage 2 consisted of one session with the original paired associations (OPA), two new paired associations (2NPA), six new paired associations (6NPA), or a new maze (NM) condition. A non-rewarded probe session followed each condition. **(A)** In the first stage, the actor–critic (pink), symbolic (blue), and neural (orange) agents demonstrated a gradual decrease in the average latency to reach six cue-specific goals and spent an increasing ratio of time at the correct target across probe sessions PS1, PS2, and PS3. **(B)** In the second stage, all agents showed above chance visit ratios for the OPA condition. However, only the symbolic and neural agents showed above chance visit ratios during the 2NPA and 6NPA conditions, demonstrating one-shot learning of two and six new paired associates. All agents demonstrated chance performance (16.7%, one out of six targets) after one session in the NM condition. **(C)** After 20 sessions of training in the OPA condition, agents were introduced to 12 new paired associates for a single trial with reward locations randomly chosen out of 43 remaining positions. The actor–critic agent learned at most one paired associate while the symbolic agent learned all 12 paired associates. Neural agents with a feedforward layer or reservoir trained with the LMS or EH rule showed a monotonic increase in one-shot learning

capacity when reservoir size was increased from 128 to 1024 units. The reservoir-based schema agent learned more paired associates than the feedforward-based agent with the same number of units. **(D)** Example trajectories during probe sessions PS1, PS2, PS3, OPA, 2NPA, 6NPA, and NM. Trajectory colors correspond to cue-specific targets. **(E)** The actor–critic agent could not learn distinct value and policy maps after a single trial for cue 7 while both symbolic and neural agents learned cue specific maps after a single trial of learning. 720 simulations were performed per agent, shaded area and error bars indicate 95% confidence interval. Asterisks **** indicates t-test against chance performance with p-value < $10^{-4}$.

## One-shot learning of multiple paired association navigation with an obstacle

We have shown that schema agents perform one-shot learning of multiple paired association navigation, while the actor–critic agent is unable to. We have also demonstrated that schema agents are unable to navigate past an obstacle to a single goal, while actor–critic agents and the hybrid actor–critic-schema agents can. Here we study one-shot learning of navigating past an obstacle to multiple goals.

The flavor-location configurations for OPA, 2NPA, and 6NPA are as in Figure 5, but with an obstacle that resembles an "H" (Fig. 6C, D). Stage 1 was extended to 50 sessions with non-rewarded probes during Session 2, 18, and 36. In Stage 2, agents were introduced to OPA, 2NPA, or 6NPA conditions for a single session followed by a probe session. The agent's starting positions were constrained so that direct heading alone could not be used to reach the goal (see Table 2).

During Stage 1, all agents except the symbolic agent gradually learned to navigate past the obstacle to reach the correct goal. By Session 50, mean latency decreased to $117 \pm 5\ s$ (actor–critic), $91 \pm 3\ s$ (neural), $141 \pm 7\ s$ (actor–critic–symbolic), $79 \pm 4\ s$ (actor–critic–neural), whereas the symbolic agent's latency increased to $343 \pm 9\ s$. Visit ratios increased monotonically for all agents except the symbolic agent to above chance performance from PS1 to PS3 (Fig. 6A). In Stage 2, only the hybrid actor–critic–symbolic and actor–critic–neural agents demonstrated one shot learning with above chance visit ratio performance after a single session in 2NPA (d=1.28 and 0.44 respectively) and in 6NPA (d=1.51 and 0.59 respectively; n=480 simulations per agent, $p<10^{-4}$). In contrast, neither the actor–critic nor the pure schema agents demonstrated one-shot learning (Fig. 6B).

The failure modes differed across agent classes. The actor–critic agent learned distinct cue-specific value and policy maps for navigating past the obstacle during Stage 1 (Fig. 6C, D), but failed to form new cue-specific maps after a single trial in Stage 2, resulting in chance performance. The symbolic agent relied exclusively on direct heading and became trapped at the obstacle (Fig. 6C). The neural schema agent avoided becoming stuck during Stage 1 due to stochasticity and decaying recall value, which introduced sufficient trajectory variability to bypass the obstacle. However, this randomness did not support one-shot learning in Stage 2, leading to chance-level performance.

Hybrid actor–critic–schema agents overcame both limitations by learning a mixed policy: state-dependent actions from the actor–critic component guided the agent away from the obstacle, while the NAVIGATE schema provided direct heading once an unobstructed path was available (Fig. 6C, D). Consequently, only the hybrid agents formed appropriate cue-specific value and policy maps for newly introduced paired associates after a single session. These results show that explicit integration of schema-based navigation with reinforcement-learning-based control is sufficient, among the agents tested, for one-shot learning of multiple paired associations in environments with obstacles, though other mechanisms may also be possible (Refer to Discussion). Performance as a function of different $\beta^{control}$ values are shown in Figure S5.

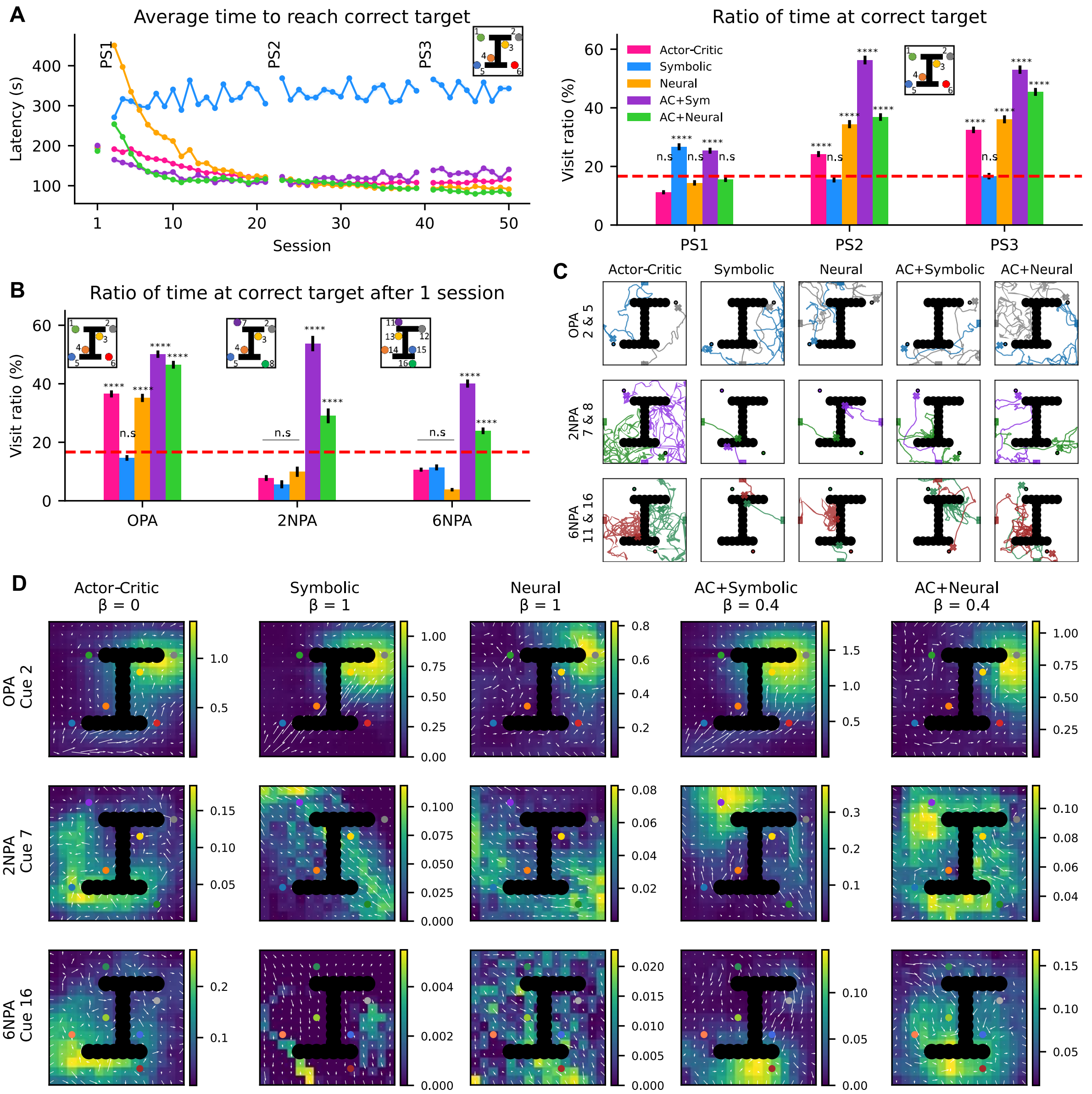

**Figure 6. One-shot navigation around an obstacle by hybrid actor–critic-schema agents. (A)** In Stage 1, all agents, except the symbolic agent, showed a decrease in average latency for the six paired associates (left) and improvements in visit ratio during the probe sessions (right). **(B)** In Stage 2, both the actor–critic–symbolic and actor–critic–neural agents demonstrated one-shot learning in the 2NPA and 6NPA conditions while the actor–critic, symbolic and neural agents showed chance performance of 16.7%. **(C)** Example trajectories for two paired associates by each agent (left to right: actor–critic, symbolic, neural, actor–critic–symbolic, actor–critic–neural) during the OPA (Cue 1 and Cue 6), 2NPA (Cue 7 and Cue 8), and 6NPA (Cue 11 and Cue 16) probe sessions after a single session in Stage 2. Although the actor–critic agent can navigate past the obstacle, it cannot navigate to the new goals after a single trial while the pure schema agents cannot navigate past the obstacle after a single trial. Only the hybrid agents use a combination of direct heading and state-based actions to navigate past the obstacle and learn new paired associates after a single trial. **(D)** Superimposed value and policy maps of all agents during non-rewarded probe sessions (averaged over 96 simulations per agent). Refer to the insets in Fig. 6B for cue-specific target locations. Only the hybrid agents show appropriate value and policy maps for the new paired associates. 480 simulations were performed per agent, shaded area and error bars indicate 95% confidence interval. Asterisk **** indicates t-test against chance performance with p-value < $10^{-4}$.

## One-shot learning with working memory and learning to gate working memory

In the previous sections, sensory cues were presented throughout the trial for all tasks. In this section, we used the same two-stage MPA task as in Figure 5 but with the sensory cue presented transiently, only at the start of the trial for 2 seconds, as was the case in the biological experiment (Tse et al., 2007). Selectively inhibiting the prefrontal cortex disrupted gradual and one-shot paired association navigation, suggesting the importance of working memory (Tse et al., 2011). Here, we wanted to study the importance of working memory in task rule learning and generalization.

We previously showed that a bump attractor can provide working memory sufficient for learning paired associates when cues are transiently presented (Kumar et al., 2022). Accordingly, we augmented the hybrid schema agent with a bump attractor receiving cue inputs and the attractor dynamics maintained cue-specific activity throughout the trial. The bump attractor output was provided as an additional input to the reservoir (Fig. 7A). However, the structure of the transient cue MPA task cannot be learned by indiscriminately maintaining sensory information in working memory. Instead, correct performance requires a form of meta-control to selectively encode the cue presented at trial onset and maintaining it while ignoring irrelevant sensory inputs, as a form of meta-control over working memory (Kumar, Manoogian, et al., 2025; Lloyd et al., 2012; O'Reilly & Frank, 2006; Todd et al., 2009). To test this, we introduced sensory distractors (Cues 17 or 18) once or twice per trial (see Methods). These distractors activated different bump-attractor subpopulation, often overwriting the target cue representation (Fig. 7B).

To enable selective working memory maintenance, we developed a neural implementation of a working-memory gating mechanism (see Methods) that regulates information flow from the sensory cue to the bump attractor (Lloyd et al., 2012; O'Reilly & Frank, 2006; Todd et al., 2009). Reservoir activity projected to two gating units governed by a winner-take-all mechanism such that only one unit multiplicatively permitted information flow from sensory input to update the bump attractor. Synapses from the reservoir to the gating units were modified using the same TD error–modulated plasticity rule as for policy learning (Eq. 45). While the winner-take-all and multiplicative gating were implemented algorithmically, the synaptic learning rule itself was biologically plausible.

Figure 7C shows the TD error (purple) and the gating activity (green, $\chi(t)$) averaged across all six cues in each probe session. After one training session (PS1), the TD error spiked during cue presentation (red bars) and remained close to zero during distractor presentation (blue bars). However, the gating activity remained near 0.45, indicating roughly equal probability of updating or maintaining working memory. After six additional training sessions (PS2), gating activity was near 0.5 during cue presentation but rapidly dropped to ~0.1 when the cue was silenced. The pattern persisted through PS3 and the OPA conditions, and critically, generalized to the 2NPA and 6NPA conditions, despite the introduction of novel cues. Thus, the reward prediction error signal drove the acquisition of a task rule that selectively updates the bump attractor during target cue presentation and maintains the representation during distractor presentation – a rule that transferred to novel paired associates.

When distractors were absent and synaptic plasticity for the gating mechanism was switched off (green), the agent successfully learned Stage 1 (Fig. 7D) and the OPA condition in Stage 2 (Fig. 7E). Average latency (Fig. 7D, top) gradually decreased to $56 \pm 3$ s by Session 20, and visit ratio improved (Fig. 7D, bottom) from PS1 to PS3 (d=3.90) and in OPA (d=4.56; n=480 simulations, $p<10^{-4}$).
Introducing distractors without gating substantially impaired performance. With one (light blue) or two distractors (dark blue), latency decreased only to $133 \pm 6$ s and $178 \pm 7$ s respectively, and visit ratios were significantly lower than the no-distractor condition (with one distractor: F=654, with two distractors: F=277; n=480 simulations, $p<10^{-4}$). Conversely, when synaptic plasticity was switched on, performance with distractors progressively improved. Latency decreased to $71 \pm 4$ s and $87 \pm 5$ s for one (yellow) and two (red) distractors, respectively, accompanied by higher visit ratios across probe sessions (with one distractor: F=2486, with two distractors: F=1670; $p<10^{-4}$). These results demonstrate successful acquisition of the task rule mediated by working memory gating mechanism.

In the second stage, one-shot learning with transient cues was possible even without gating when distractors were absent (green bars in Fig. 7E). However, when distractors were presented with no gating, visit ratios fell to chance for 2NPA (one distractor: d=0.07, two distractors: d=0.05) and below chance for 6NPA (one distractor: d=-0.39, two distractors: d=-1.09; light blue and dark blue bars in Fig. 7E), demonstrating that the bump attractor alone without gating was insufficient for one-shot learning with distractors. When synaptic plasticity for the gating mechanism was switched on, one-shot learning was rescued in both the 2NPA and 6NPA conditions despite the presence of distractors (yellow and red bars in Fig. 7E). Visit ratios increased for both 2NPA (one distractor: d=0.45, two distractors: d=0.78) and 6NPA (one distractor: d=1.46, two distractors: d=1.83; n=480 simulations, all $p<10^{-3}$), demonstrating the generalization of the learned working memory gating rule for robust one-shot learning in the presence of distractors.

Finally, we examined synaptic weight changes (Fig. 7F), which have been used to compare model predictions with biological plasticity dynamics (McClelland, 2013; Tse et al., 2011). Synaptic weight change was quantified as the mean squared difference between the synaptic weights before and after OPA, 2NPA, 6NPA, or NM session in the second stage. Plasticity at synapses onto the goal neurons exhibited a non-monotonic dependence on task novelty. Weight change increased from OPA to 2NPA to 6NPA but was larger for both 2NPA (paired d=0.18) and 6NPA (paired d=0.59) compared to NM (n=480 simulations, p<0.001). Hence, overall plasticity did not increase monotonically with novelty from OPA to NM, resembling plasticity dynamics reported in the anterior cingulate cortex (Tse et al., 2011). The reduced weight change in NM likely reflects the lower probability of successfully locating goals in the NM, which limited reward-driven goal-synapse plasticity.

In contrast, synapses onto the coordinate neurons and gating units showed a monotonic increase in plasticity with increasing novelty. Weight changes for coordinate neuron synapses (ANOVA F=9658) and gating neuron synapses (ANOVA F=33; n=480 simulations, $p<10^{-4}$) increased from OPA to NM, consistent with plasticity dynamics observed in the hippocampus (Tse et al., 2011). Although 2NPA and 6NPA produced comparable weight changes in these populations (d=0.06), both exceeded those observed in OPA and approached levels seen in NM, indicating novelty sensitivity in spatial representation and control mechanisms. Weight changes at actor and critic synapses were lowest in the OPA condition, but differences between the 2NPA, 6NPA and NM conditions were negligible in magnitude (all $|d| \leq 0.10$). Together, these plasticity patterns reveal a functional dissociation between associative memory, spatial representation, working-memory control, and policy learning mechanisms, with each exhibiting distinct sensitivity to novelty and task demands.

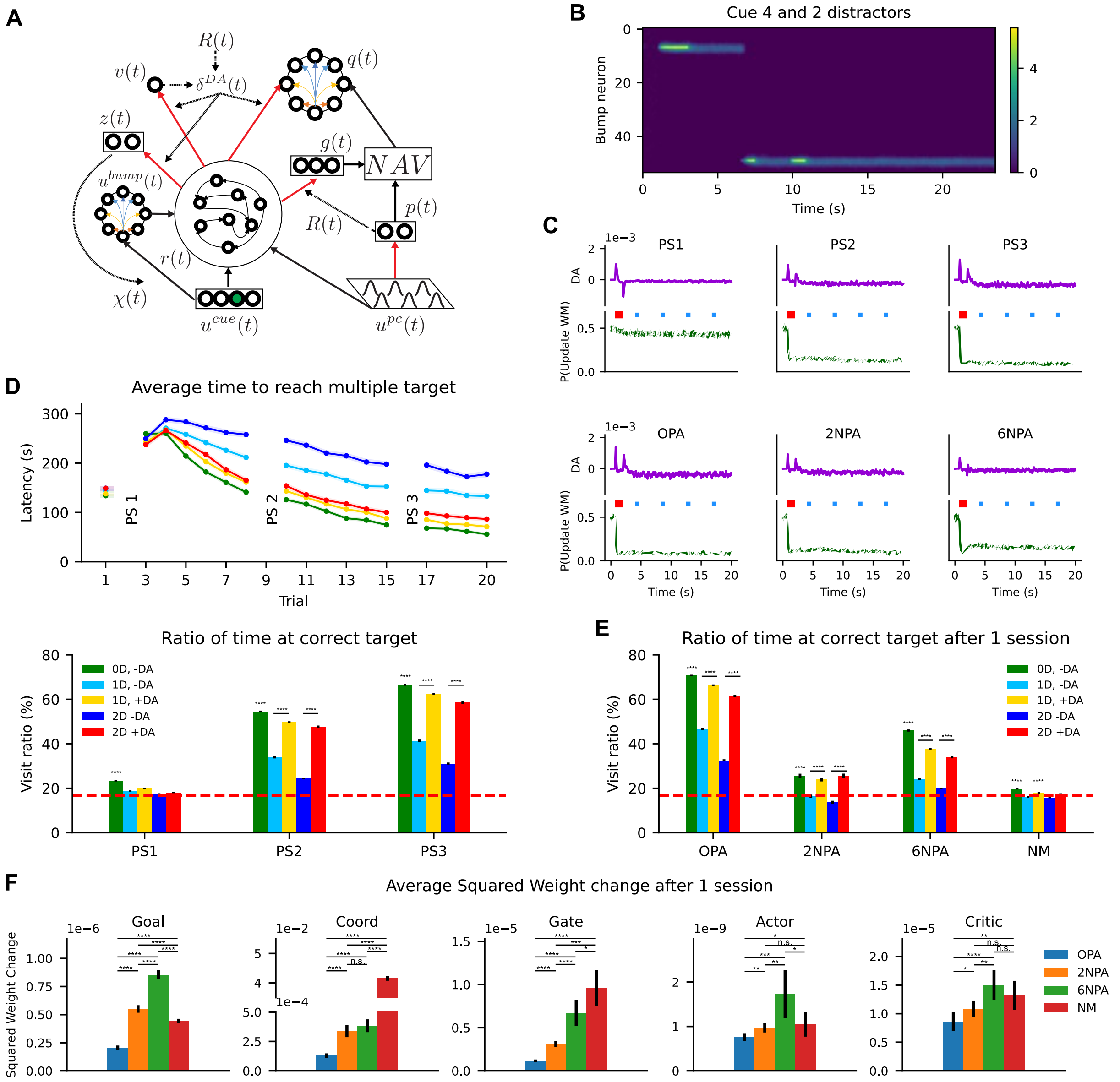

**Figure 7. Learning to gate working memory generalizes to new paired associates. (A)** A bump attractor and working memory gating mechanism were added to the actor–critic–neural agent. The synapses from the reservoir to the gating units were modified using the TD error-modulated plasticity. **(B)** The bump attractor maintains a persistent representation of a transient cue 4, but presentation of a distractor with cue 17 activates a different subpopulation. **(C)** The TD error (purple) and gating activity (green) averaged across all six cues in each probe session are shown for PS1, PS2, PS3, OPA, 2NPA, and 6 NPA conditions with plasticity of the synapses onto the gating units switched on. The gating policy gradually learned through PS1 and PS2 persisted for PS3 and OPA, and generalized to 2NPA and 6NPA even though these included new paired associates that had been only encountered in a single trial. **(D)** Latencies (top) and visit ratios (bottom) with zero (0D), one (1D), or two (2D) distractors. In the first stage of the task, as the number of distractors increased when plasticity was switched off (-DA), agents took a longer time to reach the goal (top) and spent less time at the goal (bottom). When plasticity was switched on (+DA), agents were more comparable in the time needed to reach the goal (top) and the time spent at the goal (bottom). **(E)** Agents one-shot learning ability was rescued even when distractors were present when synaptic plasticity for the gating mechanism was switched on. **(F)** Synaptic weight change as novelty increases from OPA to 2NPA to 6NPA to NM for synapses onto the goal neurons, coordinate neurons, gating units, actor, and critic. 480 simulations were performed per agent, shaded area and error bars indicate 95% confidence intervals. Asterisks indicates t-test against chance performance and a line underneath the asterisks indicates a paired t-test.

# Discussion

Rodent one-shot learning in the two-stage multiple paired association (MPA) navigation task was suggested to be schema-dependent (Tse et al., 2007). However, the conceptualization of these schemas, their role in enabling one-shot learning, and biological implementations have remained unclear. We formalized and composed candidate schemas into a biologically plausible agent that may explain rodent one-shot learning. The LEARN METRIC REPRESENTATION schema gradually acquires a low-dimensional coordinate representation for self-localization across environments. The LEARN FLAVOR–LOCATION schema enables rapid binding and deletion between sensory cues and goal coordinates using two biologically plausible synaptic plasticity rules. The NAVIGATE schema converts current and recalled goal coordinates into goal-directed actions. This decomposition provides a mechanistic account of how gradual structure learning gives rise to one-shot learning.

Although the successor representation learns the transition structure of an environment (Dayan, 1993; Stachenfeld et al., 2017), its quadratic formulation limits scalability. Grid-cell–like representations may also serve as a basis for a stable metric code, but additional circuit mechanisms are required to transform place or grid-cell-like representations into explicit goal-directed vectors (Banino et al., 2018; Bush et al., 2015; Fiete et al., 2008; Wen et al., 2024). Alternatively, LEARN METRIC REPRESENTATION provides a reusable low-dimensional substrate using place cells for self-localization and vector-based navigation across environments. However, a monotonic rate code trades resolution against environment size (Fiete et al., 2008), and reported coordinate-like tuning (LaChance et al., 2019; Lever et al., 2009) does not establish it. Hence, it would be interesting to develop alternative models that do not use coordinate neurons.

Hopfield networks have been suggested to model hippocampal CA3 (Hopfield, 1982; Pfeiffer & Foster, 2015; Rolls, 2013), however, continual learning is difficult due to catastrophic interference (Fusi, 2021; Parisi et al., 2019). Additionally, hetero-associative networks may not be able to delete erroneous associations (Chandra et al., 2025; Tyulmankov et al., 2021). LEARN FLAVOR–LOCATION supports rapid cue–goal binding using local reward-modulated plasticity and recall accuracy degrades smoothly with increasing memory load. This is because cue-specific associations are stored in separable representational subspaces, consistent with contextual gating strategies proposed to reduce interference during continual learning (Lin et al., 2023; Zeng et al., 2019). Additionally, acetylcholine modulated plasticity (Zannone et al., 2018) deletes cue-specific associations, which reduces perseveration for one-shot spatial learning.

Navigation behavior reflects sensitivity to goal distance and direction rather than stimulus–response chains (Etienne et al., 1998; Müller & Wehner, 1988). Neural recordings reveal cue-invariant goal-vector (Howard et al., 2014; Ito et al., 2015; Ormond & O'Keefe, 2022), and Cartesian or polar coordinate–like tuning (LaChance et al., 2019; Lever et al., 2009) representations in hippocampal and related circuits. These suggest vector-subtraction–like mechanisms as a biological substrate for navigation and generalization. Although our neural

NAVIGATE schema was pretrained to approximate vector subtraction, units in the penultimate layer developed distance- and angle-selective tuning that qualitatively resemble goal-vector coding in hippocampal CA1 (Høydal et al., 2019; Sarel et al., 2017). We therefore interpret NAVIGATE not as a computation learned de novo during the MPA task, but as acquired through evolution or development (Dan et al., 2024; Lyu et al., 2022; Poucet & Hok, 2017), with the latter using vectorized error signals (Francioni et al., 2026; Silva et al., 2024). However, vector-based navigation only affords direct goal heading, while flexible behavior requires integration with state-dependent action-selection mechanisms when obstacles block straight-line paths. Empirically, hippocampal goal-direction signals often remain oriented toward the goal even during detours (Sarel et al., 2017), whereas striatal value representations support route selection around barriers (van der Meer & Redish, 2009). We therefore frame flexible navigation as arising from parallel schema-based and reinforcement-learning control systems, possibly arbitrated by the anterior cingulate cortex (Akam et al., 2021; Collins, 2025; Daw et al., 2011).

Although Tse et al. (2007) did not probe dopaminergic involvement during schema-dependent learning, our model predicts that reward prediction error signals reinforce not only navigation policies (Kumar, Bordelon, et al., 2025; Yaghoubi et al., 2026) but also the working-memory gating policy that selects which sensory cues should be maintained for subsequent cue–goal binding as a form of meta-control. This implies phasic dopamine signals should reflect the acquisition of cue relevance, while prefrontal cortex should increasingly maintain task-relevant cues and suppress distractors (Kumar, Manoogian, et al., 2025; O'Reilly & Frank, 2006). Working memory gating may therefore be an additional slow-learning component that supports schema-dependent one-shot behavior under more demanding temporal contingencies.

A key requirement for biological plausibility is that task learning within the proposed schemas should be implemented by synaptic update rules depending only on local activity variables and global neuromodulatory factors, rather than non-local synapse-specific information. The plasticity rules in LEARN METRIC REPRESENTATION and LEARN FLAVOR–LOCATION satisfy this criterion. Learning a metric representation only requires an axis-specific error term and presynaptic place cell activity, yielding a local two-factor learning rule observed in the hippocampus (Katsuki et al., 1991; Nicoll & Schmitz, 2005; Urban & Barrionuevo, 1996), and can be extended to a three-factor Hebbian rule (Frémaux et al., 2013). Our proposed four-factor rule extends the canonical Hebbian plasticity rule (Frémaux & Gerstner, 2016; Hoerzer et al., 2012; Overwiening et al., 2025) with an additional reward factor. Hence, one-shot associative memory and vector-based navigation can be implemented using local synaptic update rules with global reward and novelty signals.

More generally, our results suggest that the two-stage MPA task involves a slow learning process that gradually establishes structured metric representations and meta-control to gate working memory, and a fast process that supports one-shot cue–goal association (Bernardi et al., 2020; Botvinick et al., 2019; Sun et al., 2025). While we did not formalize the gating-learning process as a computational-level schema, it involves more structure than a standard actor–critic agent while still trained by the same TD error. This suggests that schema learning and model-free reinforcement learning may not be sharply separable, paralleling debates about model-based versus model-free control (Bein & Niv, 2025; Dolan & Dayan, 2013). Schema acquisition may be better conceptualized as the gradual emergence of structured mechanisms that enable fast learning, rather than as a discrete system or switch.

Consistent with the slow–fast learning interpretation, hippocampal and prefrontal manipulations implicate these regions in both gradual schema acquisition and rapid one-shot cue–goal binding. Our model mirrors their learning dynamics: slow processes such as LEARN METRIC REPRESENTATION and working-memory gating exhibited monotonic novelty-dependent synaptic weight changes resembling hippocampal gene-expression patterns, whereas the fast LEARN FLAVOR–LOCATION process showed nonmonotonic weight changes paralleling prefrontal patterns (Tse et al., 2011). Our model has other plausible neural analogs including basal ganglia for actor–critic (Kumar et al., 2022; Niv, 2009), and prefrontal–basal ganglia–thalamic circuits for working-memory gating (Mukherjee et al., 2021; Rikhye et al., 2018). How these regions synergistically compose schema (Sun et al., 2025; Zong et al., 2025) remains an important open question.

Comparisons with deep meta-reinforcement learning (Wang et al., 2018) further support the proposed schema architecture. A meta-RL agent with symbolic LEARN FLAVOR-LOCATION achieved one-shot learning, but this disappeared when place-cell selectivity was shuffled between environments. This demonstrates that meta-RL agents learned a NAVIGATE-like computation tied to a fixed spatial representation, but not a reusable metric-representation schema. Conversely, LEARN METRIC REPRESENTATION continues to support one-shot navigation after place-cell shuffles, suggesting that reusable schema acquisition may require additional constraints (Schaeffer et al., 2022).

One limitation is that NAVIGATE supports direct heading, whereas flexible behavior may require explicit planning mechanisms for detours. Future models could incorporate model-based path-planning (Stöckl & Maass, 2023; Zhang et al., 2022) or hierarchical reinforcement learning (Botvinick, 2012; McGovern & Barto, 2001), although it remains unclear how these approaches can be biologically instantiated, while preserving flexible one-shot navigation to multiple goals (Tessereau et al., 2021). More importantly, how the proposed schema architecture explains other cognitive phenomena including hierarchical associations (McKenzie et al., 2014; Ribas-Fernandes et al., 2011), memory consolidation (Alvarez & Squire, 1994; Kumaran et al., 2016; Spens & Burgess, 2024), and other behavior and neural features of schemas (Baraduc et al., 2019; Sun et al., 2025; Zhou et al., 2020) remains open. Implementing the schema architecture with spiking neurons would enable closer comparison with experimental data (Brzosko et al., 2017; Frémaux et al., 2013).

Despite these limitations, our model makes several testable predictions about the neural mechanisms underlying schema-dependent one-shot navigation. First, the existence of neurons encoding low-dimensional coordinate-like representations (LaChance et al., 2019; Lever et al., 2009) whose activity causally supports rapid vector-based navigation. Selective disruption of such representations in the hippocampus, prefrontal cortex, subiculum or entorhinal cortex should impair one-shot navigation while leaving gradual learning relatively intact. Second, obstacle navigation should depend on parallel hippocampal and striatal control signals, with hippocampal goal-direction representations remaining oriented toward the goal during detours and striatal value signals guiding state-dependent action selection (van der Meer & Redish, 2009). Third, dopaminergic prediction-error signals should encode the temporal structure of task-relevant cues and reinforce a working-memory gating policy that enables one-shot learning despite transient cues and distractors (O'Reilly & Frank, 2006).

## Code and data availability

The code for all our models, simulations, analyses and figures is written in Python and is available in this GitHub repository https://github.com/mgkumar138/schema4one.

## Author contributions

M.G.K. performed the simulations and analyses, and drafted the paper. M.G.K., C.T., C.L., S-C.Y., and A.Y.Y.T. designed the research, discussed the findings, and revised the paper.

## Conflict of interests

The authors declare no competing interests.

## Acknowledgements

We thank Hui Min Tan and Roger Herikstad for discussions, and Adam Lee for advice on meta-reinforcement learning. This work was supported by the following grants: Singapore Ministry of Health (National Medical Research Council Open Fund-Individual Research Grants NMRC/OFIRG/0043/2017 and OFIRG25jul-0148/MOH-002096-00 to A.Y.Y.T. and MOH-000962 to S-C.Y.); National University of Singapore Young Investigator Award (NUSYIA_FY16_P22 to A.Y.Y.T.); Singapore Ministry of Education Academic Research Fund Tier 3 (MOE2017-T3-1-002 to C.L. and S-C.Y.); Singapore Ministry of Education Academic Research Fund Tier 2 (T2EP30121-0027 to C.L.); National University of Singapore Healthy Longevity Translational Research Programme Pilot Grant (HLTRP/2022/PS-04 to A.Y.Y.T.); National University of Singapore Artificial Intelligence Seed Funding (NAII-SF-2024-008 to A.Y.Y.T.) and the NUS Development Grant (to M.G.K.). The computational work for this article was performed on the High Performance Computing resources at the National University of Singapore and Harvard University.

# References

Akam, T., Rodrigues-Vaz, I., Marcelo, I., Zhang, X., Pereira, M., Oliveira, R. F., Dayan, P., & Costa, R. M. (2021). The Anterior Cingulate Cortex Predicts Future States to Mediate Model-Based Action Selection. *Neuron*, *109*(1), 149-163.e7. https://doi.org/10.1016/j.neuron.2020.10.013

Alvarez, P., & Squire, L. R. (1994). Memory consolidation and the medial temporal lobe: A simple network model. *Proceedings of the National Academy of Sciences of the United States of America*, *91*(15), 7041–7045. https://doi.org/10.1073/pnas.91.15.7041

Banino, A., Barry, C., Uria, B., Blundell, C., Lillicrap, T., Mirowski, P., Pritzel, A., Chadwick, M. J., Degris, T., Modayil, J., Wayne, G., Soyer, H., Viola, F., Zhang, B., Goroshin, R., Rabinowitz, N., Pascanu, R., Beattie, C., Petersen, S., … Kumaran, D. (2018). Vector-based navigation using grid-like representations in artificial agents. *Nature*, *557*(7705), 429–433. https://doi.org/10.1038/s41586-018-0102-6

Baraduc, P., Duhamel, J.-R., & Wirth, S. (2019). Schema cells in the macaque hippocampus. *Science*, *363*(6427), 635–639. https://doi.org/10.1126/science.aav5404

Bartlett, F. C., & Burt, C. (1932). Remembering: a Study in Experimental and Social Psychology. *British Journal of Educational Psychology*, *3*(2), 187–192. https://doi.org/10.1111/j.2044-8279.1933.tb02913.x

Bein, O., & Niv, Y. (2025). Schemas, reinforcement learning and the medial prefrontal cortex. *Nature Reviews Neuroscience 2025 26:3*, *26*(3), 141–157. https://doi.org/10.1038/s41583-024-00893-z

Bellec, G., Scherr, F., Subramoney, A., Hajek, E., Salaj, D., Legenstein, R., & Maass, W. (2020). A solution to the learning dilemma for recurrent networks of spiking neurons. *Nature Communications*, *11*(1), 1–15. https://doi.org/10.1038/s41467-020-17236-y

Bentall, R. P., Corcoran, R., Howard, R., Blackwood, N., & Kinderman, P. (2001). PERSECUTORY DELUSIONS: A REVIEW AND THEORETICAL INTEGRATION. *Clinical Psychology Review*, *21*(8), 1143–1192. https://doi.org/10.1016/S0272-7358(01)00106-4

Bernardi, S., Benna, M. K., Rigotti, M., Munuera, J., Fusi, S., & Salzman, C. D. (2020). The Geometry of Abstraction in the Hippocampus and Prefrontal Cortex. *Cell*, 954–967. https://doi.org/10.1016/j.cell.2020.09.031

Binz, M., Dasgupta, I., Jagadish, A. K., Botvinick, M., Wang, J. X., & Schulz, E. (2024). Meta-learned models of cognition. *Behavioral and Brain Sciences*, *47*, e147. https://doi.org/10.1017/S0140525X23003266

Botvinick, M. (2012). Hierarchical reinforcement learning and decision making. *Current Opinion in Neurobiology*, *22*(6), 956–962. https://doi.org/10.1016/j.conb.2012.05.008

Botvinick, M., Ritter, S., Wang, J. X., Kurth-Nelson, Z., Blundell, C., & Hassabis, D. (2019). Reinforcement Learning, Fast and Slow. *Trends in Cognitive Sciences*, *23*(5), 408–422. https://doi.org/10.1016/j.tics.2019.02.006

Brzosko, Z., Zannone, S., Schultz, W., Clopath, C., & Paulsen, O. (2017). Sequential neuromodulation of Hebbian plasticity offers mechanism for effective reward-based navigation. *ELife*, *6*, 1–18. https://doi.org/10.7554/elife.27756

Bush, D., Barry, C., Manson, D., & Burgess, N. (2015). Using Grid Cells for Navigation. *Neuron*, *87*(3), 507–520. https://doi.org/10.1016/j.neuron.2015.07.006

Chandra, S., Sharma, S., Chaudhuri, R., & Fiete, I. (2025). Episodic and associative memory from spatial scaffolds in the hippocampus. *Nature 2025 638:8051*, *638*(8051), 739–751. https://doi.org/10.1038/s41586-024-08392-y

Collins, A. G. E. (2025). A habit and working memory model as an alternative account of human reward-based learning. *Nature Human Behaviour 2025 10:2*, *10*(2), 357–369. https://doi.org/10.1038/s41562-025-02340-0

Czekalla, N., Schröder, A., Mayer, A. V, Stierand, J., Stolz, D. S., Kube, T., Korn, C. W., Wilhelm-Groch, I., Klein, J. P., Paulus, F. M., Krach, S., & Müller-Pinzler, L. (2026). Aberrant insula activity to negative and reduced learning from positive feedback underlie maladaptive self-beliefs in depression. *Transl Psychiatry,* **16**, 397 (2026). https://doi.org/10.1038/s41398-026-04341-7

Dan, C., Hulse, B. K., Kappagantula, R., Jayaraman, V., & Hermundstad, A. M. (2024). A neural circuit architecture for rapid learning in goal-directed navigation. *Neuron*, *112*(15), 2581-2599.e23. https://doi.org/10.1016/j.neuron.2024.04.036

Daw, N. D., Gershman, S. J., Seymour, B., Dayan, P., & Dolan, R. J. (2011). Model-based influences on humans' choices and striatal prediction errors. *Neuron*, *69*(6), 1204–1215. https://doi.org/10.1016/j.neuron.2011.02.027

Dayan, P. (1993). Improving Generalization for Temporal Difference Learning: The Successor Representation. *Neural Computation*, *5*(4), 613–624. https://doi.org/10.1162/neco.1993.5.4.613

Dohare, S., Hernandez-Garcia, J. F., Lan, Q., Rahman, P., Mahmood, A. R., & Sutton, R. S. (2024). Loss of plasticity in deep continual learning. *Nature 2024 632:8026*, *632*(8026), 768–774. https://doi.org/10.1038/s41586-024-07711-7

Dolan, R. J., & Dayan, P. (2013). Goals and habits in the brain. *Neuron*, *80*(2), 312–325. https://doi.org/10.1016/j.neuron.2013.09.007

Doya, K. (2000). Reinforcement learning in continuous time and space. *Neural Computation*, *12*(1), 219–245. https://doi.org/10.1162/089976600300015961

Etienne, A. S., Maurer, R., Berlie, J., Reverdin, B., Rowe, T., Georgakopoulos, J., & Séguinot, V. (1998). Navigation through vector addition. *Nature*, *396*(6707), 161–164. https://doi.org/10.1038/24151

Fiete, I. R., Burak, Y., & Brookings, T. (2008). What grid cells convey about rat location. *Journal of Neuroscience*, *28*(27), 6858–6871. https://doi.org/10.1523/JNEUROSCI.5684-07.2008

Finn, C., Abbeel, P., & Levine, S. (2017). Model-Agnostic Meta-Learning for Fast Adaptation of Deep Networks. In D. Precup & Y. W. Teh (Eds.), *Proceedings of the 34th International Conference on Machine Learning* (Vol. 70, pp. 1126--1135). PMLR. https://doi.org/10.48550/arXiv.1703.03400

Foster, D. J., Morris, R. G., & Dayan, P. (2000). A model of hippocampally dependent navigation, using the temporal difference learning rule. *Hippocampus*, *10*(1), 1–16. https://doi.org/10.1002/(SICI)1098-1063(2000)10:1<1::AID-HIPO1>3.0.CO;2-1

Francioni, V., Tang, V. D., Toloza, E. H. S., Ding, Z., Brown, N. J., & Harnett, M. T. (2026). Vectorized instructive signals in cortical dendrites. *Nature 2026*, 1–10. https://doi.org/10.1038/s41586-026-10190-7

Frémaux, N., & Gerstner, W. (2016). Neuromodulated Spike-Timing-Dependent Plasticity, and Theory of Three-Factor Learning Rules. *Frontiers in Neural Circuits*, *9*, 85. https://doi.org/10.3389/fncir.2015.00085

Frémaux, N., Sprekeler, H., & Gerstner, W. (2013). Reinforcement Learning Using a Continuous Time Actor-Critic Framework with Spiking Neurons. *PLoS Computational Biology*, *9*(4). https://doi.org/10.1371/journal.pcbi.1003024

Fusi, S. (2021). Memory capacity of neural network models. *ArXiv Preprint ArXiv:2108.07839*, 1–29. https://doi.org/https://doi.org/10.48550/arXiv.2108.07839

Gagne, C., Zika, O., Dayan, P., & Bishop, S. J. (2020). Impaired adaptation of learning to contingency volatility in internalizing psychopathology. *ELife*, *9*, 1–51. https://doi.org/10.7554/ELIFE.61387

Gilboa, A., & Marlatte, H. (2017). Neurobiology of Schemas and Schema-Mediated Memory. *Trends in Cognitive Sciences*, *21*(8), 618–631. https://doi.org/10.1016/j.tics.2017.04.013

Hochreiter, S., & Schmidhuber, J. (1997). Long Short-Term Memory. *Neural Computation*, *9*(8), 1735–1780. https://doi.org/10.1162/NECO.1997.9.8.1735

Hoerzer, G. M., Legenstein, R., & Maass, W. (2012). Emergence of complex computational structures from chaotic neural networks through reward-modulated hebbian learning. *Cerebral Cortex*, *24*(3), 677–690. https://doi.org/10.1093/cercor/bhs348

Hopfield, J. J. (1982). Neural networks and physical systems with emergent collective computational abilities. *Proceedings of the National Academy of Sciences of the United States of America*, *79*(8), 2554–2558. https://doi.org/10.1073/pnas.79.8.2554

Hospedales, T. M., Antoniou, A., Micaelli, P., & Storkey, A. J. (2021). Meta-Learning in Neural Networks: A Survey. *IEEE Transactions on Pattern Analysis and Machine Intelligence*, 1–20. https://doi.org/10.1109/TPAMI.2021.3079209

Howard, L. R., Javadi, A. H., Yu, Y., Mill, R. D., Morrison, L. C., Knight, R., Loftus, M. M., Staskute, L., & Spiers, H. J. (2014). The hippocampus and entorhinal cortex encode the path and euclidean distances to goals during navigation. *Current Biology*, *24*(12), 1331–1340. https://doi.org/10.1016/j.cub.2014.05.001

Høydal, Ø. A., Skytøen, E. R., Andersson, S. O., Moser, M. B., & Moser, E. I. (2019). Object-vector coding in the medial entorhinal cortex. *Nature*, *568*(7752), 400–404. https://doi.org/10.1038/s41586-019-1077-7

Hwu, T., & Krichmar, J. L. (2020). A neural model of schemas and memory encoding. *Biological Cybernetics*, *114*(2), 169–186. https://doi.org/10.1007/s00422-019-00808-7

Ito, H. T., Zhang, S. J., Witter, M. P., Moser, E. I., & Moser, M. B. (2015). A prefrontal-thalamo-hippocampal circuit for goal-directed spatial navigation. *Nature*, *522*(7554), 50–55. https://doi.org/10.1038/nature14396

Katsuki, H., Kaneko, S., Tajima, A., & Satoh, M. (1991). Separate mechanisms of long-term potentiation in two input systems to CA3 pyramidal neurons of rat hippocampal slices as revealed by the whole-cell patch-clamp technique. *Neuroscience Research*, *12*(3), 393–402. https://doi.org/10.1016/0168-0102(91)90070-F

Kumar, M. G., Ayyadhury, S., & Murugan, E. (2024). Trends Innovations Challenges in Employing Interdisciplinary Approaches to Biomedical Sciences. In *Translational Research in Biomedical Sciences: Recent Progress and Future Prospects* (pp. 287–308). Springer Nature Singapore. https://doi.org/10.1007/978-981-97-1777-4_20

Kumar, M. G., Bordelon, B., Zavatone-Veth, J. A., & Pehlevan, C. (2025). A Model of Place Field Reorganization During Reward Maximization. *Proceedings of Machine Learning Research*, *267*(Proceedings of the 42nd International Conference on Machine Learning), 31892–31929.

Kumar, M. G., Manoogian, A., Qian, W., Pehlevan, C., & Rhoads, S. A. (2025, November 21). Neurocomputational Underpinnings of Suboptimal Beliefs in Reinforcement Learning Agents. *8th Annual Conference on Cognitive Computational Neuroscience*. https://doi.org/10.32470/3kpf810

Kumar, M. G., Tan, C., Libedinsky, C., Yen, S.-C., & Tan, A. Y.-Y. Y. (2022). A nonlinear hidden layer enables actor–critic agents to learn multiple paired association navigation. *Cerebral Cortex*, *32*(18), 3917–3936. https://doi.org/10.1093/cercor/bhab456

Kumaran, D., Hassabis, D., & McClelland, J. L. (2016). What Learning Systems do Intelligent Agents Need? Complementary Learning Systems Theory Updated. *Trends in Cognitive Sciences*, *20*(7), 512–534. https://doi.org/10.1016/j.tics.2016.05.004

LaChance, P. A., Todd, T. P., & Taube, J. S. (2019). A sense of space in postrhinal cortex. *Science*, *365*(6449), eaax4192. https://doi.org/10.1126/science.aax4192

Lee, C., Kumar, M. G., & Tan, C. (2023). DetermiNet: A Large-Scale Diagnostic Dataset for Complex Visually-Grounded Referencing using Determiners. *International Conference on Computer Vision*, 1–8. http://arxiv.org/abs/2309.03483

Legenstein, R., Chase, S. M., Schwartz, A. B., & Maass, W. (2010). A Reward-Modulated Hebbian Learning Rule Can Explain Experimentally Observed Network Reorganization in a Brain Control Task. *Journal of Neuroscience*, *30*(25), 8400–8410. https://doi.org/10.1523/JNEUROSCI.4284-09.2010

Lever, C., Burton, S., Jeewajee, A., O'Keefe, J., & Burgess, N. (2009). Boundary Vector Cells in the Subiculum of the Hippocampal Formation. *The Journal of Neuroscience*, *29*(31), 9771–9777. https://doi.org/10.1523/JNEUROSCI.1319-09.2009

Lillicrap, T. P., Santoro, A., Marris, L., Akerman, C. J., & Hinton, G. (2020). Backpropagation and the brain. *Nature Reviews Neuroscience*, *21*(6), 335–346. https://doi.org/10.1038/s41583-020-0277-3

Lin, Z., Azaman, H., Kumar, M. G., & Tan, C. (2023). *Compositional Learning of Visually-Grounded Concepts Using Reinforcement*. 1–4. http://arxiv.org/abs/2309.04504

Lin, Z., Kumar, M. G., & Tan, C. (2025). Human-like compositional learning of visually-grounded concepts using synthetic environments. *ArXiv.Org*. https://doi.org/10.48550/ARXIV.2504.06618

Lloyd, K., Becker, N., Jones, M. W., & Bogacz, R. (2012). Learning to use working memory: a reinforcement learning gating model of rule acquisition in rats. *Frontiers in Computational Neuroscience*, *6*(October), 1–10. https://doi.org/10.3389/fncom.2012.00087

Lyu, C., Abbott, L. F., & Maimon, G. (2022). Building an allocentric travelling direction signal via vector computation. *Nature*, *601*(7891), 92–97. https://doi.org/10.1038/s41586-021-04067-0

Maass, W., Joshi, P., & Sontag, E. D. (2007). Computational aspects of feedback in neural circuits. *PLoS Computational Biology*, *3*(1), 0015–0034. https://doi.org/10.1371/journal.pcbi.0020165

Marr, D. (2010). Vision: A computational investigation into the human representation and processing of visual information. In *MIT Press* (Vol. 80, Number 4). MIT press. https://doi.org/10.7551/mitpress/9780262514620.001.0001

Marr, D., & Poggio, T. (1977). From understanding computation to understanding neural circuitry. *Neurosciences Research Program Bulletin*, *15*, 470–488.

McClelland, J. L. (2013). Incorporating rapid neocortical learning of new schema-consistent information into complementary learning systems theory. *Journal of Experimental Psychology: General*, *142*(4), 1190–1210. https://doi.org/10.1037/a0033812

McGovern, A., & Barto, A. G. (2001). Automatic Discovery of Subgoals in Reinforcement Learning using Diverse Density. *Proceedings of the Eighteenth International Conference on Machine Learning*, 361–368. https://doi.org/10.5555/645530.655681

McKenzie, S., Frank, A. J., Kinsky, N. R., Porter, B., Rivière, P. D., & Eichenbaum, H. (2014). Hippocampal representation of related and opposing memories develop within distinct, hierarchically organized neural schemas. *Neuron*, *83*(1), 202–215. https://doi.org/10.1016/j.neuron.2014.05.019

Mnih, V., Badia, A. P., Mirza, M., Graves, A., Lillicrap, T. P., Harley, T., Silver, D., & Kavukcuoglu, K. (2016). Asynchronous Methods for Deep Reinforcement Learning. *Proceedings of the 33rd International Conference on Machine Learning*, *48*, 1928–1937. https://doi.org/10.1177/0956797613514093

Movellan, J. R. (1991). Contrastive Hebbian Learning in the Continuous Hopfield Model. *Connectionist Models*, 10–17. https://doi.org/10.1016/b978-1-4832-1448-1.50007-x

Mukherjee, A., Lam, N. H., Wimmer, R. D., & Halassa, M. M. (2021). Thalamic circuits for independent control of prefrontal signal and noise. *Nature*, *600*(7887), 100–104. https://doi.org/10.1038/s41586-021-04056-3

Müller, M., & Wehner, R. (1988). Path integration in desert ants, Cataglyphis fortis . *Proceedings of the National Academy of Sciences*, *85*(14), 5287–5290. https://doi.org/10.1073/pnas.85.14.5287

Murray, J. M. (2019). Local online learning in recurrent networks with random feedback. *ELife*, *8*, 1–25. https://doi.org/10.7554/eLife.43299

Nicoll, R. A., & Schmitz, D. (2005). Synaptic plasticity at hippocampal mossy fibre synapses. *Nature Reviews Neuroscience*, *6*(11), 863–876. https://doi.org/10.1038/nrn1786

Niv, Y. (2009). Reinforcement learning in the brain. *Journal of Mathematical Psychology*, *53*(3), 139–154. https://doi.org/10.1016/j.jmp.2008.12.005

O'Reilly, R. C., & Frank, M. J. (2006). Making working memory work: A computational model of learning in the prefrontal cortex and basal ganglia. In *Neural Computation* (Vol. 18, Number 2). https://doi.org/10.1162/089976606775093909

Ormond, J., & O'Keefe, J. (2022). Hippocampal place cells have goal-oriented vector fields during navigation. *Nature*, *607*(7920), 741–746. https://doi.org/10.1038/s41586-022-04913-9

Overwiening, J., Kumar, M. G., & Sompolinsky, H. (2025). TeDFA-$\delta$: Temporal integration in deep spiking networks trained with feedback alignment improves policy learning. *8th Annual Conference on Cognitive Computational Neuroscience* . https://2025.ccneuro.org/poster/?id=Sxs0948e5w

Parisi, G. I., Kemker, R., Part, J. L., Kanan, C., & Wermter, S. (2019). Continual lifelong learning with neural networks: A review. *Neural Networks*, *113*, 54–71. https://doi.org/10.1016/j.neunet.2019.01.012

Parthasarathy, A., Tang, C., Herikstad, R., Cheong, L. F., Yen, S.-C., & Libedinsky, C. (2019). Time-invariant working memory representations in the presence of code-morphing in the lateral prefrontal cortex. *Nature Communications*, *10*(1), 4995. https://doi.org/10.1038/s41467-019-12841-y

Pfeiffer, B. E., & Foster, D. J. (2015). Autoassociative dynamics in the generation of sequences of hippocampal place cells. *Science*, *349*(6244), 180–183. https://doi.org/10.1126/science.aaa9633

Poggio, T. (2012). The levels of understanding framework, revised. *Perception*, *41*(9), 1017–1023. https://doi.org/10.1068/p7299

Poucet, B., & Hok, V. (2017). Remembering goal locations. *Current Opinion in Behavioral Sciences*, *17*, 51–56. https://doi.org/10.1016/j.cobeha.2017.06.003

Qian, W., Zavatone-Veth, J. A., Ruben, B. S., & Pehlevan, C. (2024). Partial observation can induce mechanistic mismatches in data-constrained models of neural dynamics. *Advances in Neural Information Processing Systems*, *37*, 67467–67510. https://doi.org/10.52202/079017-2154

Ribas-Fernandes, J. J. F., Solway, A., Diuk, C., McGuire, J. T., Barto, A. G., Niv, Y., & Botvinick, M. M. (2011). A Neural Signature of Hierarchical Reinforcement Learning. *Neuron*, *71*(2), 370–379. https://doi.org/10.1016/j.neuron.2011.05.042

Rikhye, R. V., Gilra, A., & Halassa, M. M. (2018). Thalamic regulation of switching between cortical representations enables cognitive flexibility. *Nature Neuroscience*, *21*(12), 1753–1763. https://doi.org/10.1038/s41593-018-0269-z

Rolls, E. T. (2013). The mechanisms for pattern completion and pattern separation in the hippocampus. *Frontiers in Systems Neuroscience*, *7*(OCT), 1–21. https://doi.org/10.3389/fnsys.2013.00074

Rumelhart, D. E. (1980). Schemata: The building blocks of cognition. In *Theoretical Issues in Reading Comprehension* (pp. 33–58). Routledge.

Rumelhart, D. E., Hinton, G. E., & Williams, R. J. (1986). Learning representations by back-propagating errors. *Nature*, *323*(6088), 533–536. https://doi.org/10.1038/323533a0

Rumelhart, D. E., & Ortony, A. (1977). The Representation of Knowledge in Memory. *Schooling and the Acquisition of Knowledge*, (January 1977), 99–135. https://doi.org/10.4324/9781315271644-10

Sarel, A., Finkelstein, A., Las, L., & Ulanovsky, N. (2017). Vectorial representation of spatial goals in the hippocampus of bats. *Science*, *355*(6321), 176–180. https://doi.org/10.1126/science.aak9589

Schaeffer, R., Khona, M., & Fiete, I. R. (2022). No Free Lunch from Deep Learning in Neuroscience: A Case Study through Models of the Entorhinal-Hippocampal Circuit. *Advances in Neural Information Processing Systems*, *35*, 16052–16067.

Sharma, S., Chandra, S., & Fiete, I. R. (2022). Content addressable memory without catastrophic forgetting by heteroassociation with a fixed scaffold. *ICML*. http://arxiv.org/abs/2202.00159

Silva, N. T., Ramírez-Buriticá, J., Pritchett, D. L., & Carey, M. R. (2024). Climbing fibers provide essential instructive signals for associative learning. *Nature Neuroscience |*, *27*, 940–951. https://doi.org/10.1038/s41593-024-01594-7

Spens, E., & Burgess, N. (2024). A generative model of memory construction and consolidation. *Nature Human Behaviour 2024 8:3*, *8*(3), 526–543. https://doi.org/10.1038/s41562-023-01799-z

Stachenfeld, K. L., Botvinick, M. M., & Gershman, S. J. (2017). The hippocampus as a predictive map. *Nature Neuroscience*, *20*(11), 1643–1653. https://doi.org/10.1038/nn.4650

Steele, R. J., & Morris, R. G. M. (1999). Delay-dependent impairment of a matching-to-place task with chronic and intrahippocampal infusion of the NMDA-antagonist D-AP5. *Hippocampus*, *9*(2), 118–136. https://doi.org/10.1002/(SICI)1098-1063(1999)9:2<118::AID-HIPO4>3.0.CO;2-8

Stöckl, C., & Maass, W. (2023). *Local prediction-learning in high-dimensional spaces enables neural networks to plan*. 1–34. https://doi.org/10.1101/2022.10.17.512572

Sun, W., Winnubst, J., Natrajan, M., Lai, C., Kajikawa, K., Bast, A., Michaelos, M., Gattoni, R., Stringer, C., Flickinger, D., Fitzgerald, J. E., & Spruston, N. (2025). Learning produces an orthogonalized state machine in the hippocampus. *Nature 2025 640:8057*, *640*(8057), 165–175. https://doi.org/10.1038/s41586-024-08548-w

Sussillo, D., & Abbott, L. F. (2009). Generating Coherent Patterns of Activity from Chaotic Neural Networks. *Neuron*, *63*(4), 544–557. https://doi.org/10.1016/j.neuron.2009.07.018

Sutton, R. S., & Barto, A. G. (2020). *Reinforcement learning: An introduction*. MIT Press.

Tessereau, C., O'Dea, R., Coombes, S., & Bast, T. (2021). Reinforcement learning approaches to hippocampus-dependent flexible spatial navigation. *Brain and Neuroscience Advances*, *5*. https://doi.org/10.1177/2398212820975634

Todd, M. T., Niv, Y., & Cohen, J. D. (2009). Learning to use working memory in partially observable environments through dopaminergic reinforcement. *Advances in Neural Information Processing Systems 21 - Proceedings of the 2008 Conference*, 1689–1696.

Tse, D., Langston, R. F., Kakeyama, M., Bethus, I., Spooner, P. a, Wood, E. R., Witter, M. P., & Morris, R. G. M. (2007). Schemas and Memory Consolidation. *Science*, *316*(5821), 76–82. https://doi.org/10.1126/science.1135935

Tse, D., Takeuchi, T., Kakeyama, M., Kajii, Y., Okuno, H., Tohyama, C., & Morris, R. G. M. (2011). Schema-Dependent Gene Activation. *Science*, *891*(August), 891–896. https://doi.org/10.1126/science.1205274

Tyulmankov, D., Fang, C., Vadaparty, A., & Yang, G. R. (2021). *Biological learning in key-value memory networks*. (NeurIPS), 1–12. http://arxiv.org/abs/2110.13976

Urban, N. N., & Barrionuevo, G. (1996). Induction of Hebbian and Non-Hebbian Mossy Fiber Long-Term Potentiation by Distinct Patterns of High-Frequency Stimulation. *The Journal*

*of Neuroscience*, *16*(13), 4293–4299. https://doi.org/10.1523/JNEUROSCI.16-13-04293.1996

van der Meer, M. A. A., & Redish, A. D. (2009). Covert expectation-of-reward in rat ventral striatum at decision points. *Frontiers in Integrative Neuroscience*, *3*(FEB), 458. https://doi.org/10.3389/neuro.07.001.2009

Wang, J., Jagadeesh, V. S., Kumar, M. G., Libedinsky, C., Yen, S.-C., Tan, A. Y.-Y., & Polepalli, J. S. (2025, December 2). Unsupervised Hebbian Learning Drives Biologically Interpretable Pattern Separation in a Hippocampal–Striatal Network. *AAAI 2026 Workshop on NeuroAI*. https://openreview.net/forum?id=1sBi8a6Cia

Wang, J. X., Kurth-Nelson, Z., Kumaran, D., Tirumala, D., Soyer, H., Leibo, J. Z., Hassabis, D., & Botvinick, M. (2018). Prefrontal cortex as a meta-reinforcement learning system. *Nature Neuroscience*, *21*(6), 860–868. https://doi.org/10.1038/s41593-018-0147-8

Wen, J. H., Sorscher, B., Aery Jones, E. A., Ganguli, S., & Giocomo, L. M. (2024). One-shot entorhinal maps enable flexible navigation in novel environments. *Nature 2024 635:8040*, *635*(8040), 943–950. https://doi.org/10.1038/s41586-024-08034-3

Whittington, J. C. R., Muller, T. H., Mark, S., Chen, G., Barry, C., Burgess, N., & Behrens, T. E. J. (2020). The Tolman-Eichenbaum Machine: Unifying Space and Relational Memory through Generalization in the Hippocampal Formation. *Cell*, *183*(5), 1249-1263.e23. https://doi.org/10.1016/j.cell.2020.10.024

Wimmer, K., Nykamp, D. Q., Constantinidis, C., & Compte, A. (2014). Bump attractor dynamics in prefrontal cortex explains behavioral precision in spatial working memory. *Nature Publishing Group*, *17*(3), 431–439. https://doi.org/10.1038/nn.3645

Yaghoubi, M., Kumar, M. G., Nieto-Posadas, A., Mosser, C. A., Gisiger, T., Wilson, É., Pehlevan, C., Williams, S., & Brandon, M. P. (2026). Predictive coding of reward in the hippocampus. *Nature 2026 651:8105*, *651*(8105), 414–420. https://doi.org/10.1038/s41586-025-09958-0

Zannone, S., Brzosko, Z., Paulsen, O., & Clopath, C. (2018). Acetylcholine-modulated plasticity in reward-driven navigation: a computational study. *Scientific Reports*, *8*(1), 9486. https://doi.org/10.1038/s41598-018-27393-2

Zeng, G., Chen, Y., Cui, B., & Yu, S. (2019). Continual learning of context-dependent processing in neural networks. *Nature Machine Intelligence*, *1*(8), 364–372. https://doi.org/10.1038/s42256-019-0080-x

Zhang, T., Rosenberg, M., Perona, P., & Meister, M. (2022). *Endotaxis: A neuromorphic algorithm for mapping, goal-learning, navigation, and patrolling*. 1–41. https://doi.org/10.1101/2021.09.24.461751

Zhou, J., Jia, C., Montesinos-Cartagena, M., Gardner, M. P. H., Zong, W., & Schoenbaum, G. (2020). Evolving schema representations in orbitofrontal ensembles during learning. *Nature*, *590*(March 2020). https://doi.org/10.1038/s41586-020-03061-2

Zong, W., Zhou, J., Gardner, M. P. H., Zhang, Z., Costa, K. M., & Schoenbaum, G. (2025). Hippocampal output suppresses orbitofrontal cortex schema cell formation. *Nature Neuroscience 2025 28:5*, *28*(5), 1048–1060. https://doi.org/10.1038/s41593-025-01928-z

# Supplementary Figures

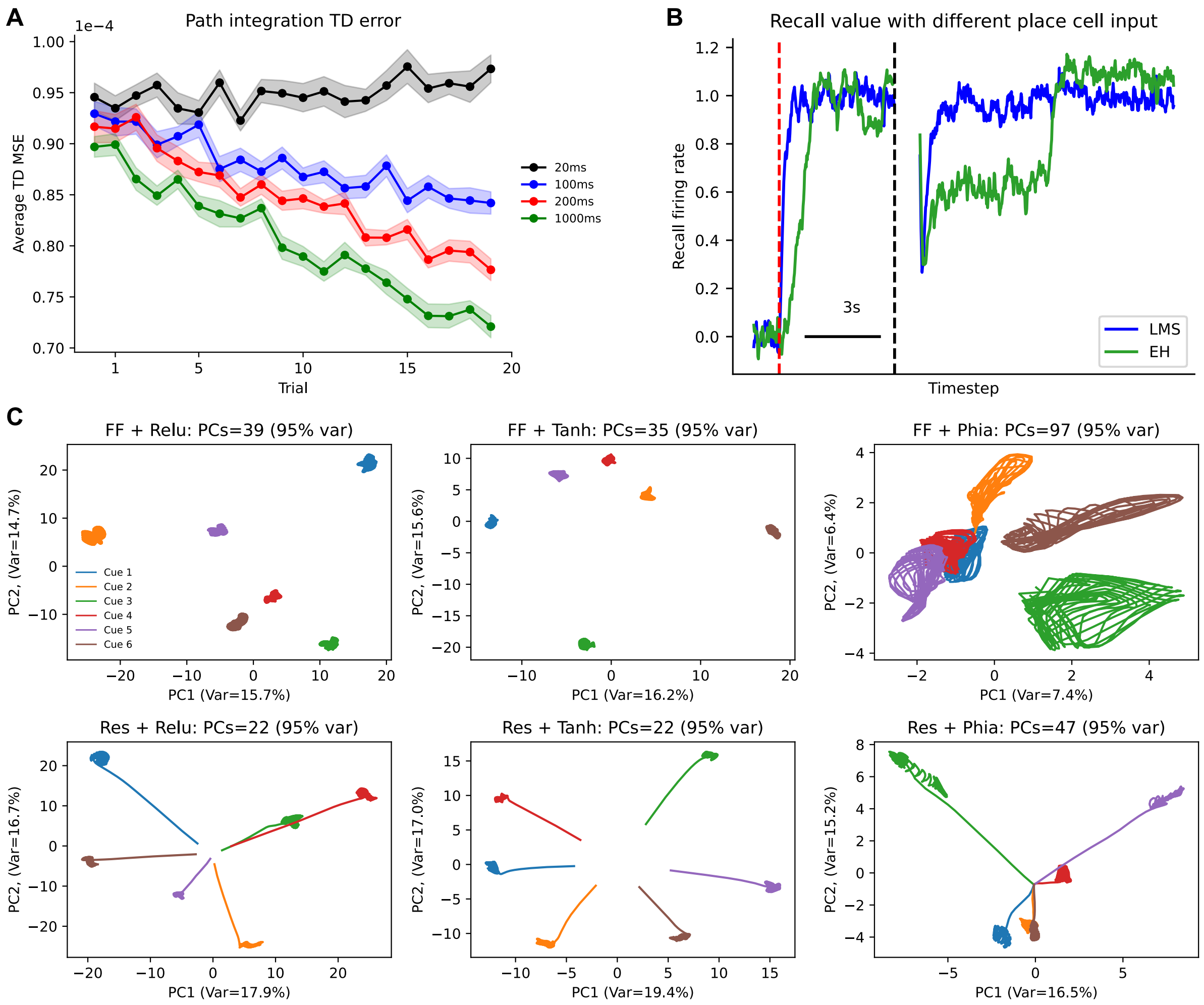

**Figure S1. Learning LEARN METRIC REPRESENTATION and FLAVOR-LOCATION schema variables. (A)** Path integration based temporal difference error (Foster et al., 2000) decreases over 20 trials as an agent performs random foraging in an open arena for 300 seconds per trial. A larger eligibility trace time constant leads to a faster decrease in TD error. **(B)** During navigation, the reservoir receives both sensory cue and place cell activity as input. When an agent receives a reward, plasticity is switched on and the agent's movement is restricted to associate the agent's current coordinates with the reservoir activity. During this period, place cell activity remains constant as the agent is static in the maze. When plasticity is switched on (red to black dashed lines), the recall value reaches 1 when either the LMS or EH rule are used. However, in the following trial when the agent is moving around the arena, causing the place cell activity to change, the recall value for the synapses trained using the LMS rule averages around 0.95 whereas the recall value for the synapses trained by the EH rule averages around 0.61. When the agent moves towards the goal, place cell activity becomes increasingly similar to when association occurred and the recalled reward value increases to 1.08. Hence, the reservoir trained using the EH rule activates the NAVIGATE schema when the agent gets closer to the goal while the reservoir trained using the LMS rule performs direct heading to achieve similar performance as the symbolic agent. **(C)** Population activity in feedforward (FF; top row) and recurrent reservoir (Res; bottom row) when the same place-cell trajectory was paired with different sensory cues and given as input to the network. For each activation function (ReLU, tanh, and $\phi_A$), activity across all spatial locations and cues was projected onto the first two principal components while retaining 95% of the variance. In feedforward networks, cue identity shifts similar spatial trajectories into distinct subspaces with separation magnitude depending on the nonlinearity, and sparse $\phi_A$producing higher-dimensional embeddings. Reservoir networks further amplify cue-dependent separation through recurrent dynamics, yielding more compact and consistently segregated cue-specific trajectories across nonlinearities. Thus, for each architecture and activation function, sensory cues selectively route spatial representations into distinct subspaces, enabling cue-specific goal associations or policies (Kumar et al., 2022) to be stored with reduced interference.

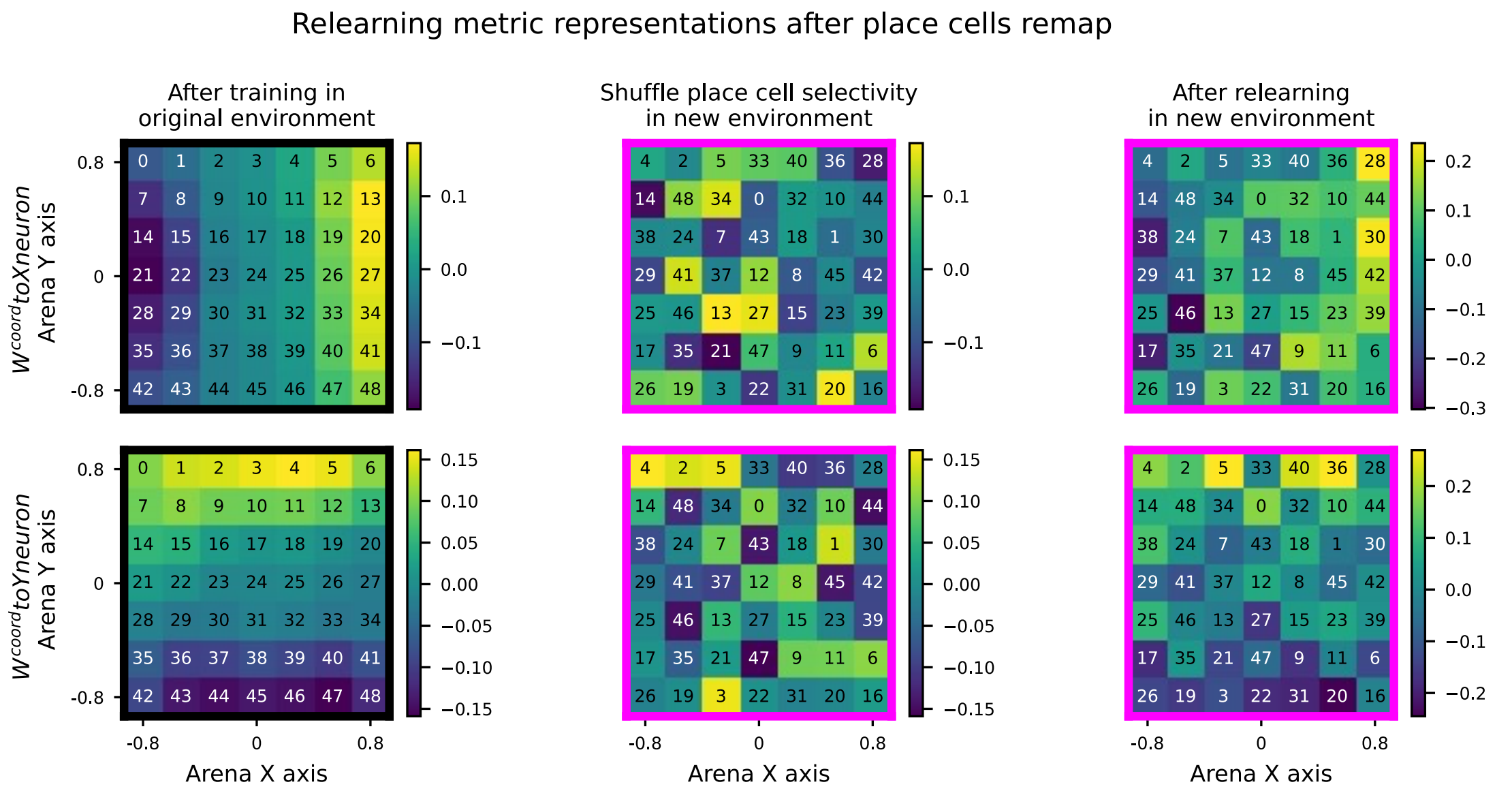

**Figure S2. Relearning X–Y metric representation after place cells remap in a new environment.** An agent only with the LEARN METRIC REPRESENTATION schema and with synaptic plasticity turned on explored a square arena using a random policy over 20 trials for 300 seconds each. All synaptic weights were initialized to zeros before learning in the original environment (not shown). The square matrices in the top and bottom rows show the synaptic strength from place cells indexed 0 to 48 to the X and Y coordinate cells respectively. **(Left)** After 20 sessions of random foraging in the original environment, the metric representation from the place cell population to the X (top) and Y (bottom) coordinate cells converged to represent the X axis and Y axis respectively for the agent to estimate its coordinates as described in Fig. 2A. 14 place cells with the most negative synaptic weights in the original environment are highlighted using white indices. **(Middle)** To model place cell remapping when an animal is placed in a new environment, we randomly shuffled place cell selectivity such that place cell 0 which had peak firing activity at coordinate (-0.8, 0.8) in the original environment demonstrates peak firing activity at coordinate (0,0.6) in the new environment. Place cell remapping leads to inaccurate coordinate estimation as the metric representation used for path integration becomes distorted. This demonstrates why the prior learned metric representation in the original environment cannot be transferred to the new environment, and a new metric representation needs to be relearned. **(Right)** As the agent explores the new environment over another 20 trials for 300 seconds each, the path integration Temporal Difference error modulated Hebbian rule gradually modifies the synaptic weights. The synaptic strength of the place cells closer to the right and left border (top row) increase and decrease, respectively to correspond to the X axis metric representation in the leftmost column. Similarly, the synaptic strength of the place cells closer to the top and bottom border (bottom row) increase and decrease, respectively to correspond to the Y axis metric representation in the leftmost column. 14 place cells with the most negative synaptic weights in the new environment are re-highlighted using white indices to demonstrate modification of synaptic weights based on the LEARN METRIC REPRESENTATION schema.

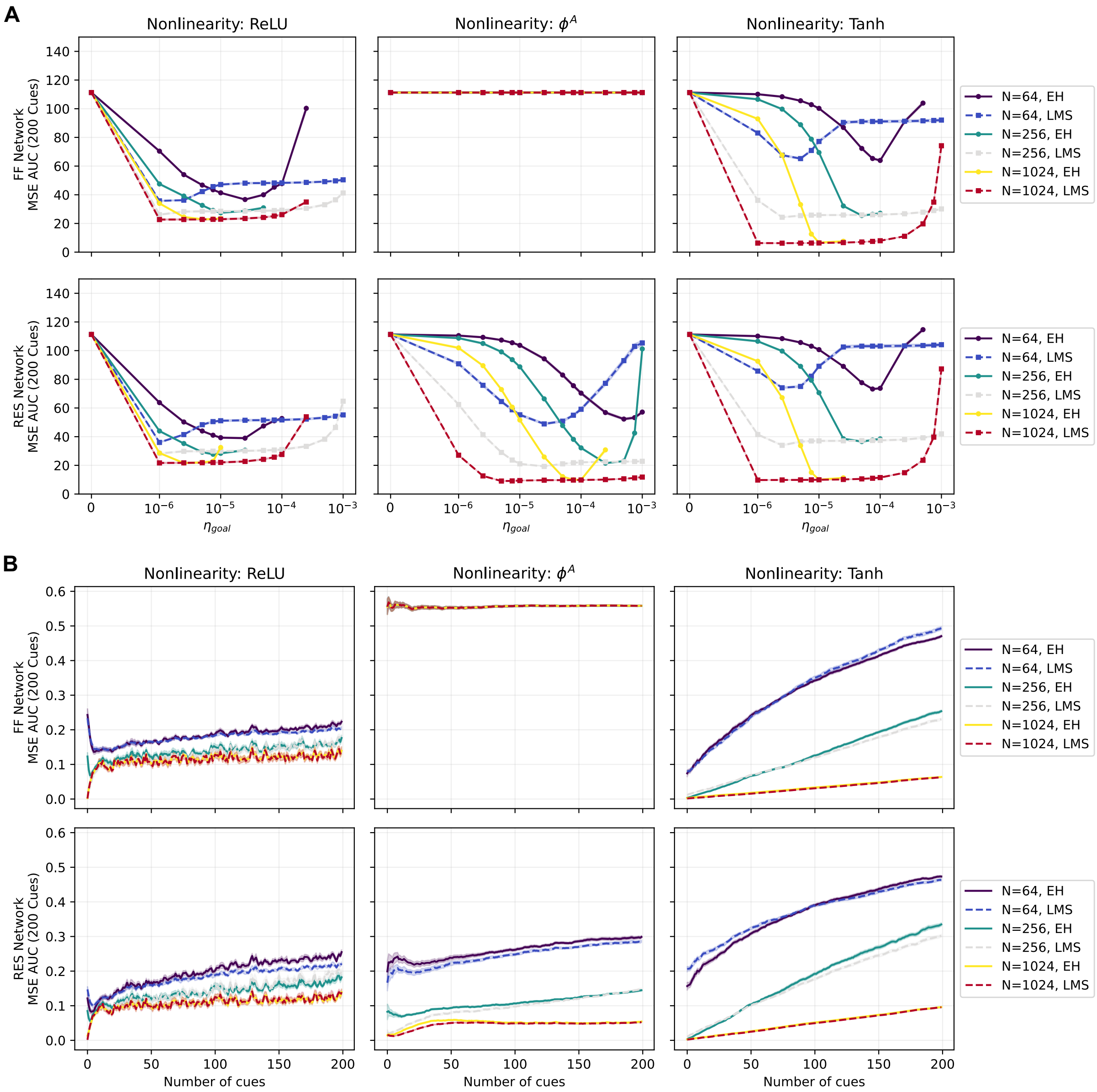

**Figure S3. Recall MSE after one-shot association of multiple cue-coordinate paired associations.** The LEARN FLAVOR-LOCATION association network receives cue vector as input and passes the activity to either a feedforward (FF) or reservoir (Res) using synapses drawn from a random uniform distribution. The reservoir uses either the Rectified Linear Unit (ReLU), the $\Phi^A$ function from Kumar et al. (2022) or the hyperbolic tangent (tanh) activation function. The output are three readout units corresponding to X, Y coordinates and recall value of 1. The synapses are modified either using the LMS or EH rule. The intent is for this network to perform similarly to a key-value matrix where after forming the cue-coordinate association, when the same cue is passed to the network, the cue specific coordinates can be recalled with a low mean squared error (MSE). **(A)** Learning rate hyperparameter sweeps show that different network size, nonlinearities and learning rules influence the choice of the optimal learning rate $\eta_{goal}$. Increasing the network size reliably decreases recall MSE for both LMS and EH rules. **(B)** Each line plot was obtained by using the optimal learning rate found by taking the value that gave the lowest area under the curve of the MSE over 200 cues. All networks show a systematic increase in recall MSE as more cues are added. The $\Phi^A$ nonlinearity causes a high degree of sparsity preventing feedforward networks from learning any meaningful

associations, whereas the reservoir network's activity allows the network to learn meaningful representations and achieve the lowest MSE compared to the tanh and ReLU nonlinearity when the size of the network is increased. When the size of the feedforward or reservoir is increased, the recall MSE is lower. With the sparse $\phi^A$ or ReLU nonlinearity, early associations yield high MSE because sparse hidden activity and weak initial readout weights limit effective credit assignment by the exploratory Hebbian rule; as additional associations are learned, increased readout synaptic magnitude enables reliable learning and lowers MSE. Line and shaded area are mean and standard error for 30 seeds.

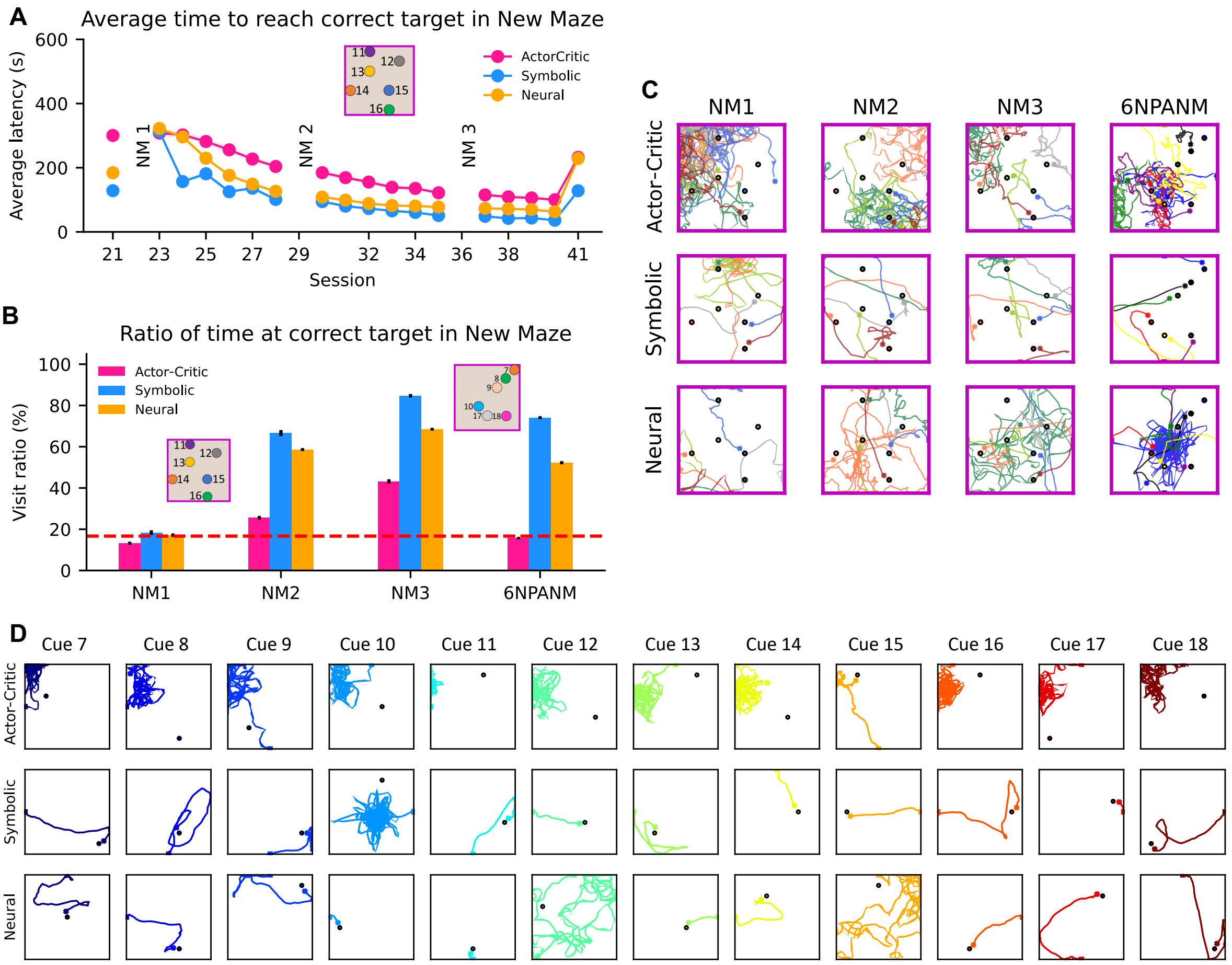

**Figure S4. One-shot navigation after place cell shuffle in a new environment (NM). (A)** After agents were trained on the MPA task for 20 sessions, they were introduced to the New Maze (NM) condition for another 20 sessions followed by one session of six new paired associations in the new maze (6NPANM). The NM condition was modelled where place cell selectivity was randomly shuffled so that the same place cells were now selective to a different location of the same arena and the cue-coordinate pairs followed the 6NPA condition. The actor–critic, symbolic, and neural agents gradually learn NM. **(B)** All agents demonstrate increasing visit ratio performances during the probe sessions NM1, NM2, and NM3. After the agents were introduced to six new paired associations in the new maze for one session, in the subsequent probe session, the actor–critic demonstrated chance performance while the symbolic and neural agents demonstrated above chance visit ratio performance. This demonstrates the continued generalizability of the schema agents where they require several trials to gradually learn a metric representation of the new environment to demonstrate one-shot learning of new paired associations. **(C)** Example trajectories of the actor–critic, symbolic, and neural agents during the probe sessions NM1, NM2, NM3, and 6NPANM. The actor–critic struggles to learn the new paired associates, whereas the symbolic agent navigates to all six paired associates. The neural agent navigates to five paired associates but fails to learn one PA (blue). **(D)** Example trajectories of the three agents during a probe session after learning 12 new paired associates for a single trial each from Fig. 5. The actor–critic navigates to the same location in the arena despite different paired associates (Cue 15). The symbolic agent fails to learn one PA (Cue 10) while the neural agent fails to learn two paired associates (Cue 12 and 15) out of 12 paired associates after a single session.

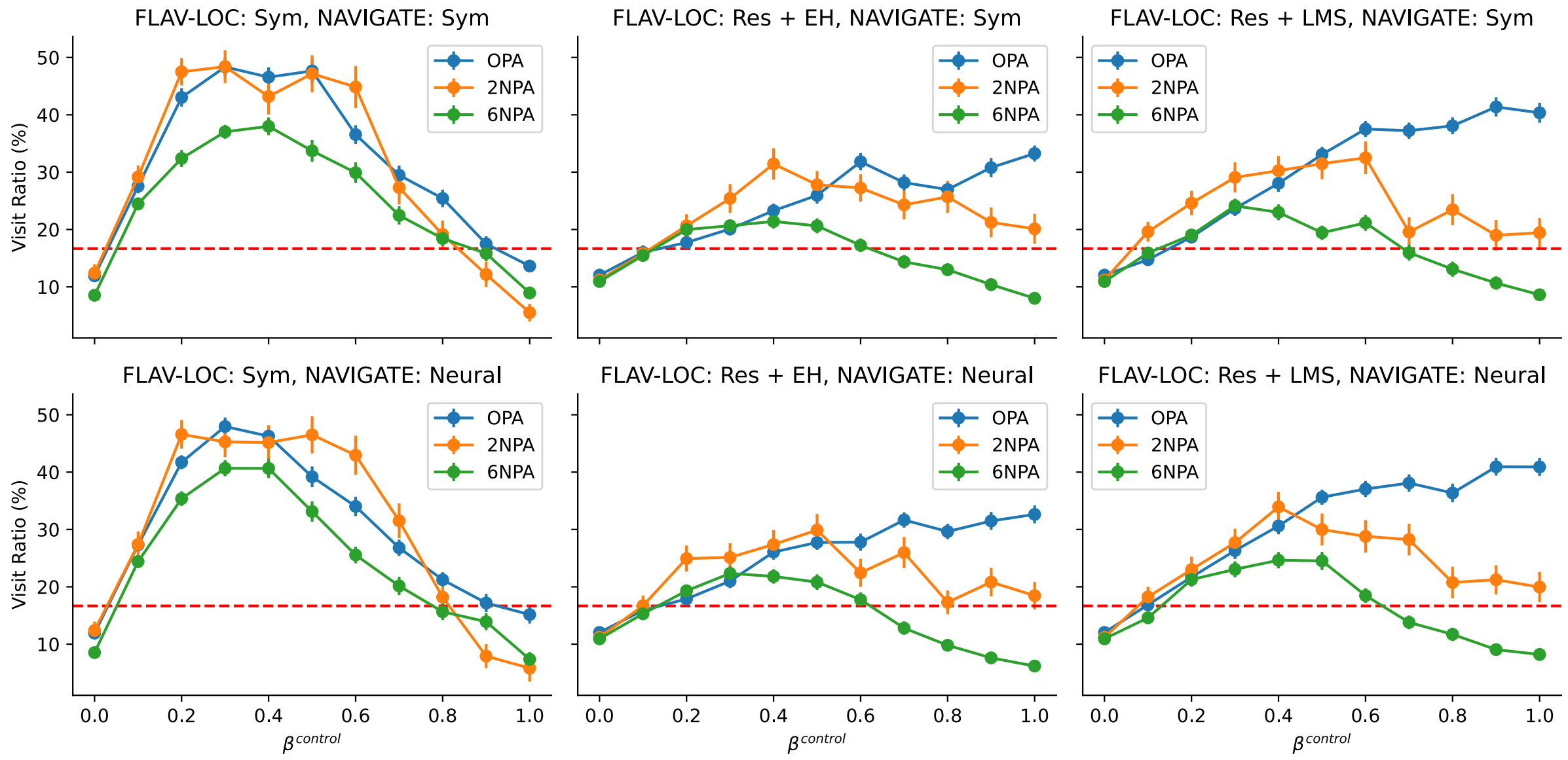

**Figure S5. Hybrid schema-actor–critic performance on MPA with obstacle for different schema implementations and learning rules.** Hybrid schema–actor–critic agents were first trained on the MPA task with obstacle for 50 sessions, then trained for 1 session on either the Original Paired Associations (OPA), 2 New Paired Associations (2NPA), or 6 New Paired Associations (6NPA) variant, followed by a non-rewarded probe session. The y-axis shows the visit ratio (percentage of time spent at the correct reward location during the probe), with chance performance at 16.7% (red horizontal dashed line). The x-axis shows $\beta^{control}$, which determines the relative influence of the schema versus actor–critic controller on action selection ($\beta^{control} = 0$: pure actor–critic; $\beta^{control} = 1$: pure schema; intermediate values: a linear summation of both controllers; see Methods for details). The six panels correspond to all combinations of NAVIGATE schema (Symbolic vs Neural) and LEARN FLAVOR—LOCATION (FLAV-LOC) association memory (symbolic key–value vs neural reservoir memory trained with either the LMS or more biologically plausible EH learning rule). Across conditions, Symbolic and Neural NAVIGATE schemas show equivalent learning performance on OPA, 2NPA, and 6NPA, even in the presence of an obstacle, with 2NPA and 6NPA explicitly demonstrating one-shot learning. However, the symbolic key–value FLAV-LOC association has higher fidelity than the neural LEARN FLAVOR–LOCATION schema. Within the neural implementations, the LMS learning rule yields slightly better performance than the more biologically plausible EH rule. Error bars indicate the standard error over 100 random seeds.

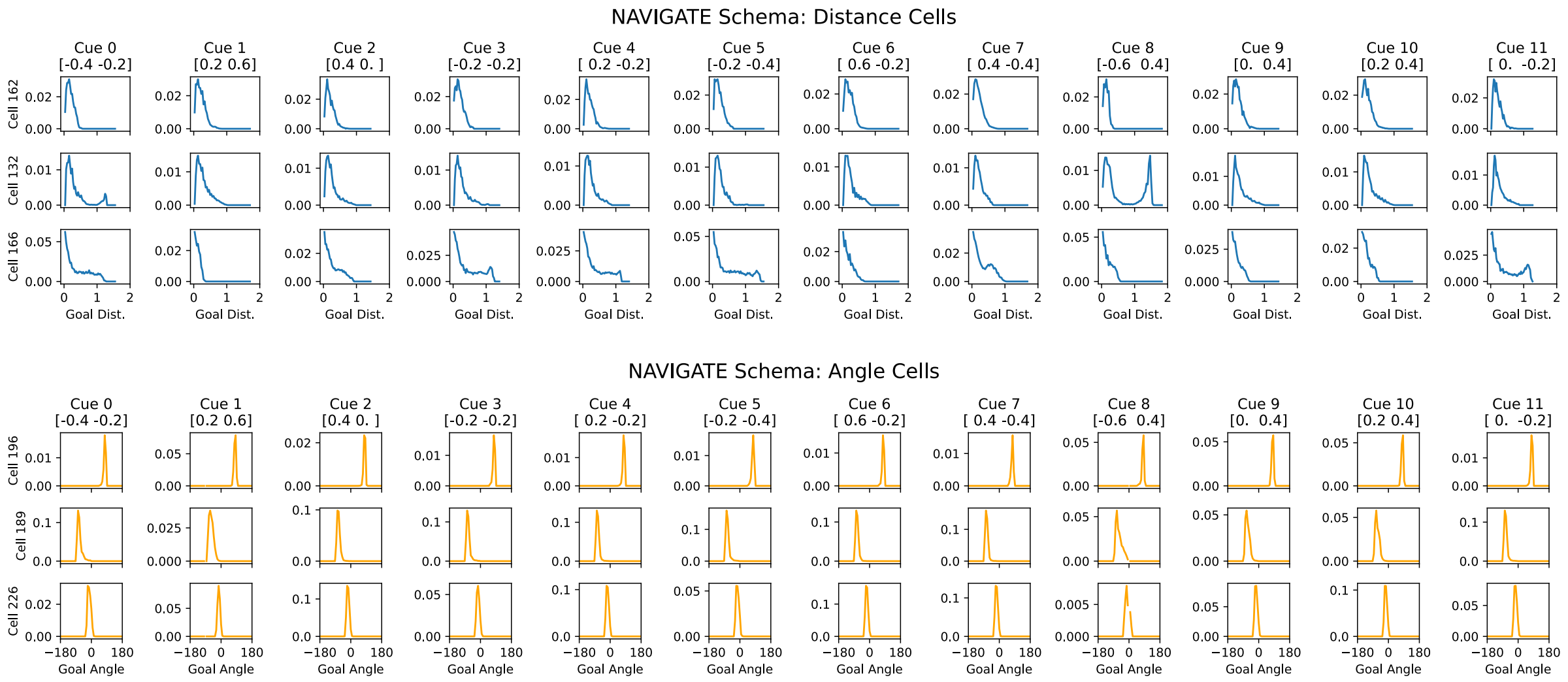

**Figure S6. Cue-invariant goal-vector coding in NAVIGATE schema.** Example units from the penultimate layer of the NAVIGATE schema network exhibit tuning consistent with distance and angle coding to the goal location. (**Top)** Distance cells show largely cue-invariant responses across multiple cue–goal associations, with firing rate varying systematically as a function of goal distance. Notably, activity changes monotonically with distance to the goal, indicating a continuous code for proximity independent of cue identity. (**Bottom)** Angle cells exhibit consistent orientation tuning across cues, with peak firing occurring at a preferred goal-bearing angle and similar tuning profiles across different cue–goal pairs. Together, these results demonstrate that NAVIGATE learns cue-invariant representations of goal distance and direction that can support vector-based navigation, similar to the hippocampal code in Sarel et al. (2017).

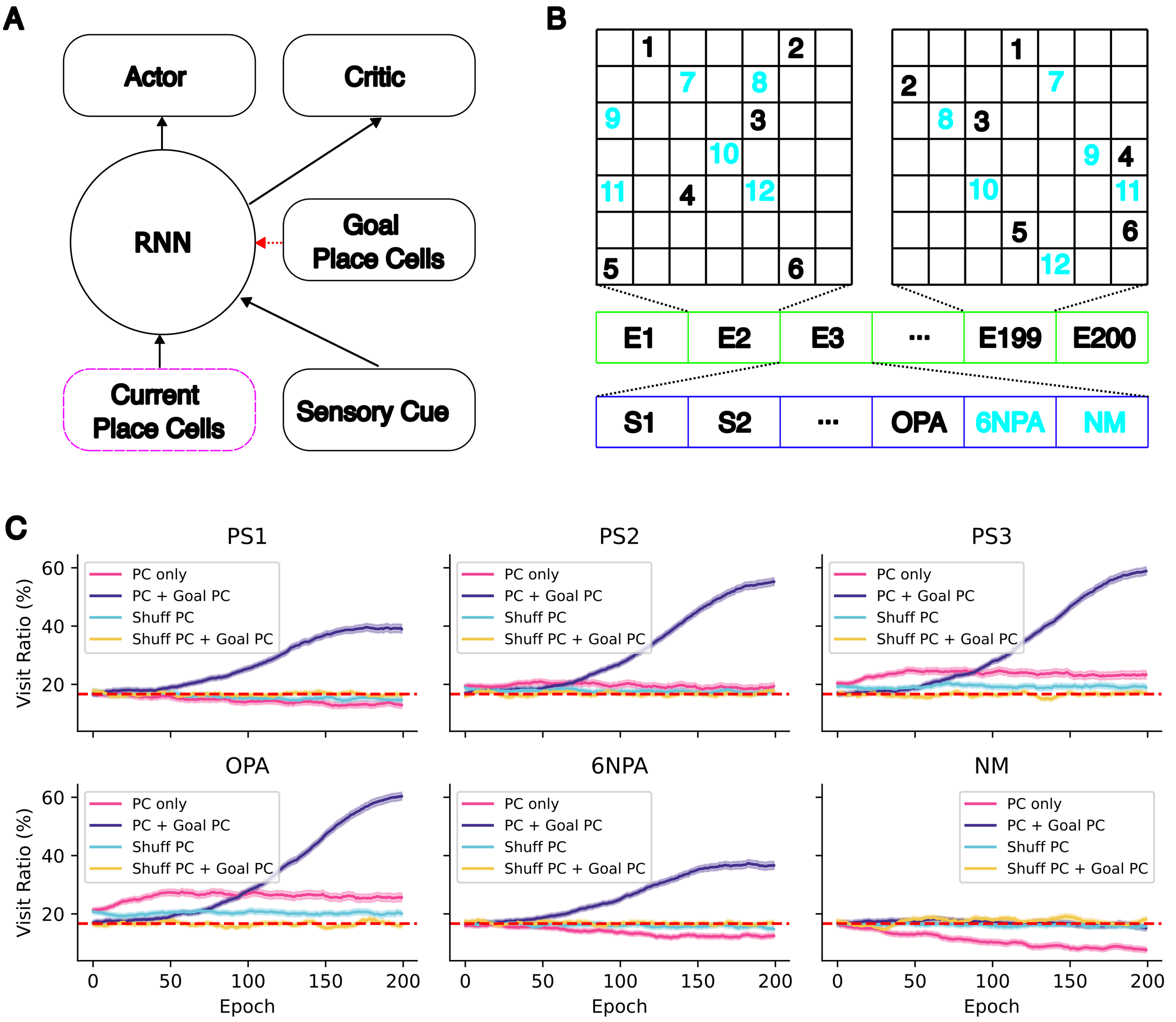

**Figure S7. Meta-reinforcement learning acquires a NAVIGATE-like computation but fails to learn a transferable metric representation schema. (A)** Meta-RL agent architecture. The agent receives a one-hot cue vector and spatial inputs (place-cell activity), and optionally a symbolic key–value memory matrix that stores cue–goal associations by providing goal place-cell activity as an additional input, similar to Banino et al. (2018) who provided goal grid cell activity as additional input, although goal grid cells have not been biologically observed. Network weights are trained using the Advantage Actor–Critic (A2C) algorithm across training epochs. To test generalization across environments, place-cell selectivity was either held fixed or shuffled between epochs to simulate remapping in a novel environment. **(B)** Inner-loop and outer-loop training structure. Within each inner-loop (blue table), the agent experiences 20 paired-association sessions, followed by probe sessions corresponding to PS1, PS2, PS3, OPA, 6NPA, or NM. In the outer-loop epochs (green table, 200 epochs), reward locations are shuffled to enforce continual adaptation. **(C)** Behavioral performance across training. Only the meta-RL agent equipped with the symbolic key–value association matrix and trained with *non-shuffled* place cells exhibits rapid emergence of one-shot learning, showing increased visit ratios during probe sessions. In contrast, shuffling place-cell selectivity across epochs abolishes one-shot learning despite continued gradual improvement on PS1 to OPA. These results suggest that meta-RL learns a NAVIGATE-like mapping from current and goal representations to actions, but does not acquire a reusable LEARN METRIC REPRESENTATION schema capable of adapting to new spatial remappings. Shaded area indicates mean and standard error

over 30 seeds. Similar learning performance was observed with five seeds for 500 and 1000 epochs.

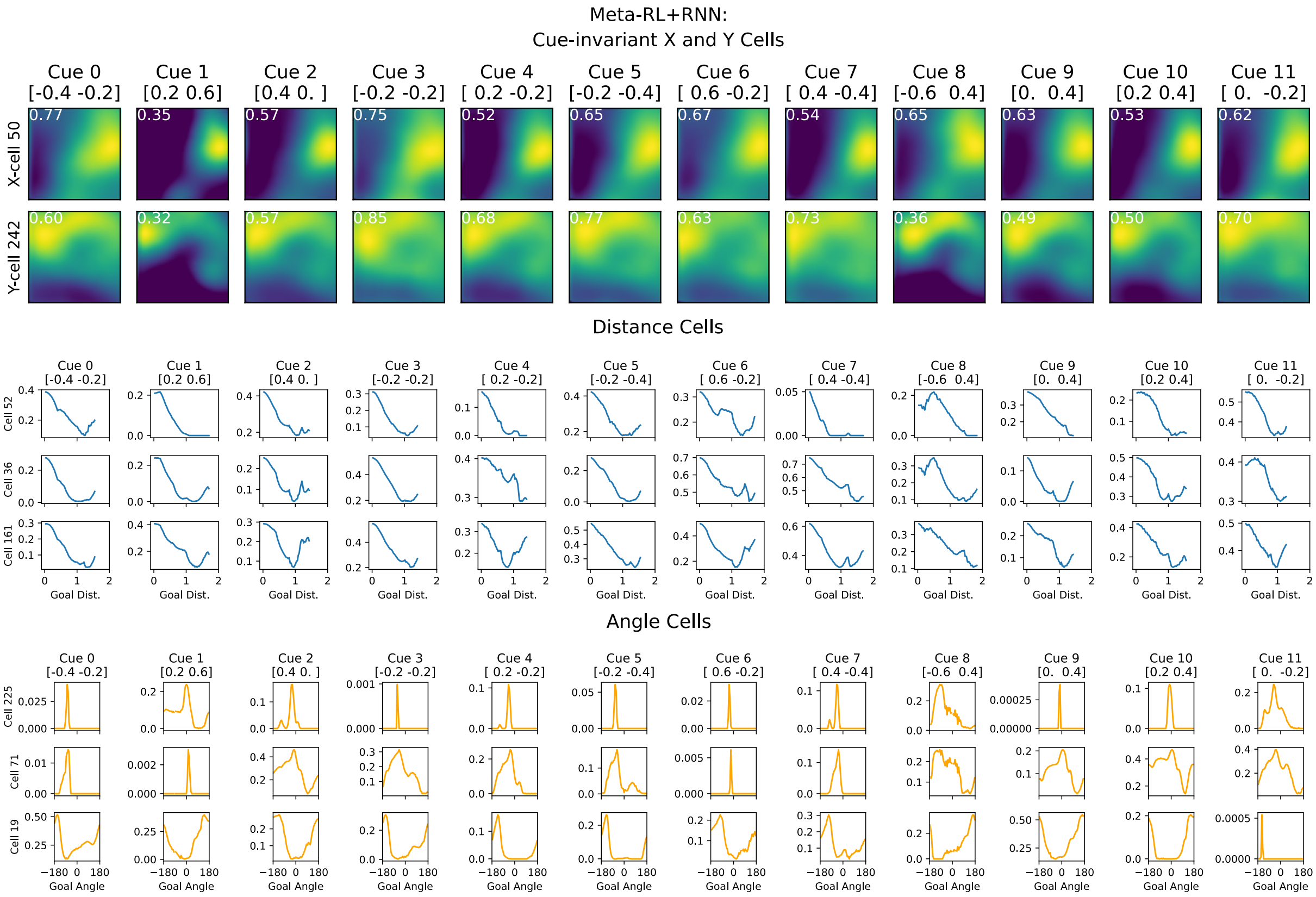

**Figure S8. Meta-RL agent learns approximate cue-invariant coordinate- and goal-vector-like representations under fixed place-cell selectivity.** Neural activity from units in the recurrent network of the meta-RL agent with a symbolic key–value memory matrix implementing LEARN FLAVOR–LOCATION was analyzed similarly to the NAVIGATE schema in Figure S6. **(Top)** Example units exhibit monotonic firing-rate activity as the agent traverses the environment along the horizontal or vertical axis, qualitatively resembling X- and Y-coordinate neurons in the LEARN METRIC REPRESENTATION schema. **(Middle)** Example units show distance-like tuning, with firing rates that increase monotonically as the agent approaches the goal, consistent with a goal-distance code. **(Bottom)** Angle-like units exhibit directional tuning relative to the goal, resembling goal-angle coding observed in the NAVIGATE schema. Across these populations, tuning profiles are partially cue-invariant, suggesting that the meta-RL agent acquires internal representations supporting vector-based navigation. However, these representations are less stable and less consistently cue-invariant across goal conditions than those produced by the explicit schema-based implementation. Notably, place-cell selectivity was held fixed across MPA environments, indicating that these coordinate- and goal-vector-like representations do not constitute a transferable LEARN METRIC REPRESENTATION mechanism capable of relearning when place cells remap across environments. Instead, consistent with behavioral results in Figure S7, the meta-RL agent appears to primarily acquire a NAVIGATE-like computation enabling vector-based navigation, similar to prior meta-RL navigation models (Banino et al., 2018), but driven here by place-cell inputs rather than grid-cell representations. Whether the one-shot learning performance deteriorates if these cells are silenced needs to be further validated.